\documentclass[journal,twoside,web]{ieeecolor}
\usepackage{jsen}
\usepackage{cite}
\usepackage{amsmath,amssymb,amsfonts}
\usepackage{algorithmic}
\usepackage{graphicx}
\usepackage{textcomp}
\usepackage{wrapfig}
\def\BibTeX{{\rm B\kern-.05em{\sc i\kern-.025em b}\kern-.08em
    T\kern-.1667em\lower.7ex\hbox{E}\kern-.125emX}}
\definecolor{abstractbg}{rgb}{0.89804,0.94510,0.83137}
\usepackage{adjustbox}
\PassOptionsToPackage{hyphens}{url}\usepackage{hyperref}
\usepackage{siunitx}
\usepackage{caption}
\usepackage{subcaption}
\usepackage{physics}
\usepackage{hyperref}

\usepackage{tikz}
\usetikzlibrary{shapes,math,arrows,fit}
\usepackage{pgfplots}
\pgfplotsset{compat=newest}
\tikzset{font=\small}
  
\definecolor{tuda_blue}{RGB}{     0, 156, 218} 
\definecolor{tuda_green}{RGB}{   80, 182, 149} 
\definecolor{tuda_yellow}{RGB}{ 255, 224,  92} 
\definecolor{tuda_red}{RGB}{    233,  80,  62} 
\definecolor{tuda_darkred}{RGB}{185,  15,  34} 

\usepackage{makecell}%

\newcommand{\figref}[1]{Fig.\,\ref{#1}}

\newcommand{\tabref}[1]{Tab.\,\ref{#1}}

\newcommand{\PAR}[1]{\vspace{-0.2eM}\vskip4pt \noindent{\bf #1}}

\makeatletter
\newcommand\notsotiny{\@setfontsize\notsotiny{7}{8}}
\makeatother
\usepackage{upgreek}
\usepackage[acronyms,toc,nonumberlist,nogroupskip]{glossaries}
\newglossary{symbols}{sym}{sbl}{List of Symbols}
\glsnoexpandfields
\loadglsentries{gloss}
\newacronym{fzd}{FZD}{Fahrzeugtechnik Darmstadt}
\newacronym{ACC}{ACC}{Adaptive Cruise Control}
\newacronym{RadarAcronym}{radar}{Radio Detection and Ranging}

\newacronym{osi}{OSI}{Open Simulation Interface~\cite{hanke_open_2017}}
\newacronym{fmi}{FMI}{Functional Mock-up Interface~\cite{modelica_association_functional_2017}}
\newacronym[plural=FMUs]{fmu}{FMU}{Functional Mock-up Unit~\cite[p.\,6]{modelica_association_functional_2017}}
\newacronym{ol}{OL}{object list}
\newacronym{pcl}{PCL}{point cloud}
\newacronym{gt}{GT}{ground truth}
\newacronym{rmse}{RMSE}{root mean squared error}
\newacronym{had}{HAD}{higly automated driving}
\newacronym{opa}{OPA}{optical phased arrays}
\newacronym{mems}{MEMS}{micro-electro-mechanical systems}
\newacronym{brdf}{BRDF}{Bidirectional Reflectance Distribution Function}
\newacronym{bsdf}{BSDF}{Bidirectional Scattering Distribution Functions~\cite{nicodemus_geometrical_1977}}
\newacronym{adc}{ADC}{analogue-digital-converter}
\newacronym{snr}{SNR}{signal to noise ratio}
\newacronym{if}{\textit{\texttt{IF}}}{interface}
\newacronym{rtk}{RTK}{real time kinematic}
\newacronym{gnss}{GNSS}{global navigation satellite system}
\newacronym{iso}{ISO}{International Organization for Standardization}
\newacronym[plural=BBs,firstplural=bounding boxes (BBs)]{bb}{BB}{bounding box}
\newacronym{fov}{FOV}{field of view}
\newacronym{iou}{IoU}{intersection over union}
\newacronym{gpu}{GPU}{graphics card}
\newacronym{6lm}{6lm}{6 Layer Model~\cite{scholtes_6-layer_2021}}
\newacronym{stf}{STF}{SeeingThroughFog~\cite{bijelic_seeing_2020}} 

\usepackage{fancyhdr}

\newcommand{\copyrightheader}{
    \footnotesize This work has been submitted to the IEEE for possible publication. Copyright may be transferred without notice, after which this version may no longer be accessible.
}

\begin{document}
\title{Simulating Road Spray Effects in Automotive Lidar Sensor Models}
\author{Clemens Linnhoff, Dominik Scheuble, Mario Bijelic, Lukas Elster, Philipp Rosenberger, Werner Ritter, Dengxin Dai and Hermann Winner
\thanks{Clemens Linnhoff, Lukas Elster, Philipp Rosenberger and Hermann Winner are with the Institute of Automotive Engineering,
Technical University of Darmstadt, 64287 Darmstadt, Germany
{\tt\small <prename>.<surname>@tu-darmstadt.de}}
\thanks{Dominik Scheuble and Werner Ritter are with the Mercedes-Benz AG, 70546 Stuttgart, Germany {\tt\small <prename>.<surname>@mercedes-benz.com}}
\thanks{Mario Bijelic is with Princeton University, 08542 Princeton, USA {\tt\small mario.bijelic@princeton.edu}}
\thanks{Dengxin Dai is with MPI for Informatics, 66123 Saarbrücken, Germany {\tt\small ddai@mpi-inf.mpg.de}}
\thanks{The authors would like to thank Max Schumacher and Patrick Pintscher for their support with conducting the experiments.}
}

\IEEEtitleabstractindextext{%
\fcolorbox{abstractbg}{abstractbg}{%
\begin{minipage}{\textwidth}%
 \begin{wrapfigure}[12]{r}{3in}%
 \vspace{-1em}
 \includegraphics[width=2.8in]{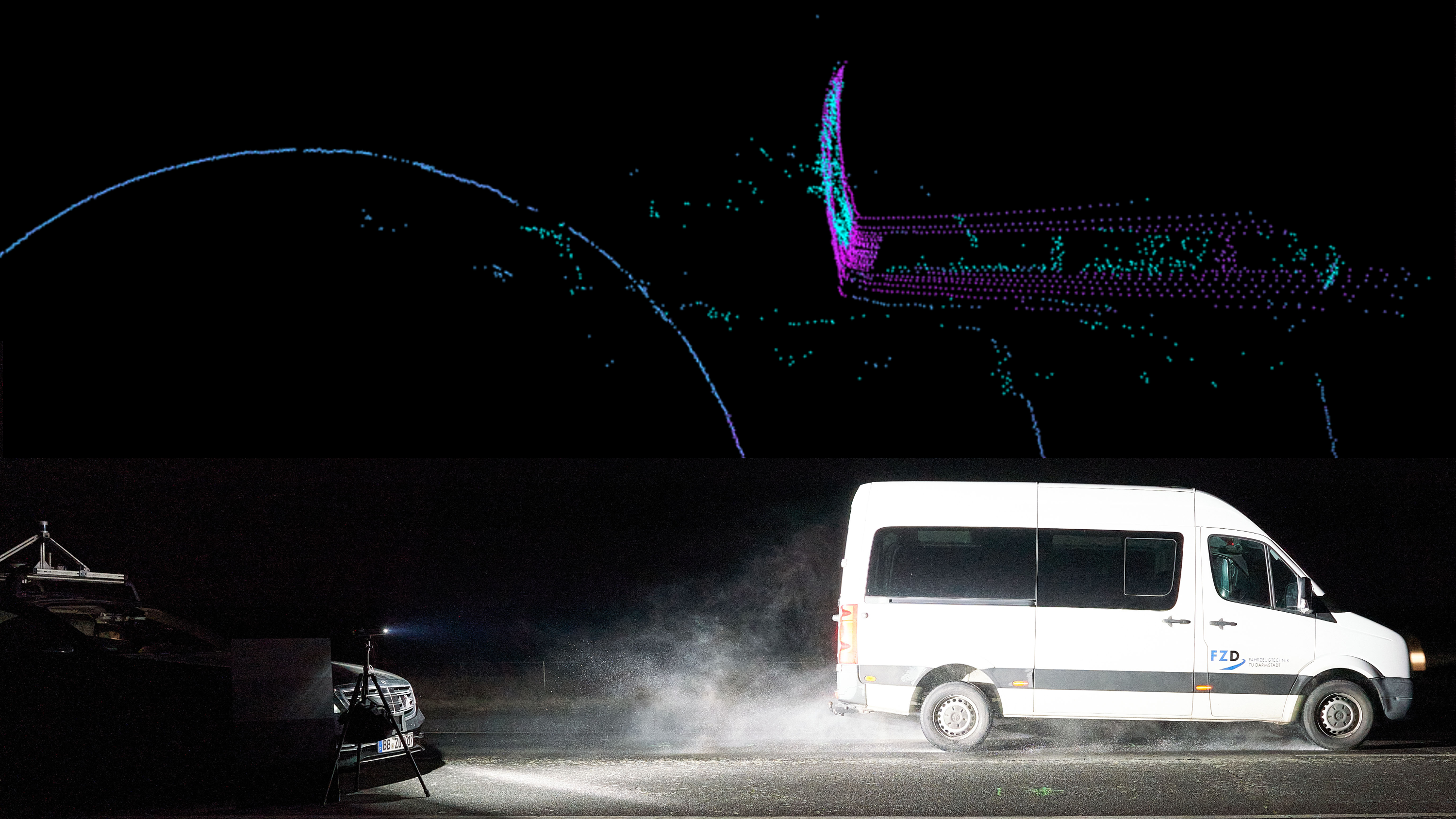}%
 \end{wrapfigure}%
\begin{abstract}
    Modeling perception sensors is key for simulation based testing of automated driving functions.
Beyond weather conditions themselves, sensors are also subjected to object dependent environmental influences like tire spray caused by vehicles moving on wet pavement.
In this work, a novel modeling approach for spray in lidar data is introduced.
The model conforms to the Open Simulation Interface (OSI) standard and is based on the formation of detection clusters within a spray plume.
The detections are rendered with a simple custom ray casting algorithm without the need of a fluid dynamics simulation or physics engine.
The model is subsequently used to generate training data for object detection algorithms.
It is shown that the model helps to improve detection in real-world spray scenarios significantly.
Furthermore, a systematic real-world data set is recorded and published for analysis, model calibration and validation of spray effects in active perception sensors.
Experiments are conducted on a test track by driving over artificially watered pavement with varying vehicle speeds, vehicle types and levels of pavement wetness.
All models and data of this work are available open source.

\end{abstract}

\end{minipage}}}

\maketitle

\thispagestyle{fancy}
\pagestyle{fancy} 
\fancyhf{} 
\fancyhead[L]{} 
\fancyhead[C]{\vtop{\vskip -20pt \centering \textcolor{gray}{\copyrightheader}}} 
\fancyhead[R]{} 
\renewcommand{\headrulewidth}{0pt} 

\section{Introduction}

Deploying automated driving systems in everyday situations requires extensive testing of each functional building block in order to ensure a safe operation under all circumstances. Due to the scarcity and hard-to-collect rare edge cases in real-world data, the replication of each functional block in simulation is necessary \cite{Dosovitskiy17carla}. Therefore, over the past years virtual testing played an increasing role in the development and validation of automated vehicles. The physical world has to be replicated accurately including material properties, object dynamics as trajectories and realistic street scenes as the replication of whole operational areas. Thereby, the degree of realism is a trade-off against computation time. Hence, simulation of perception sensors such as camera, radar and lidar requiring an approximation of the whole physical world are most challenging. This causes a dominant domain gap leading to lower results when using simulated data for training computer vision algorithms compared to real world recordings \cite{Richter_2016_ECCVGTA}. 
Currently, those restrictions arise from the insufficient replication of real-world physical effects. This hinders the deployment of automated driving systems in the most challenging surroundings, such as complex illumination conditions including nighttime driving and various weather effects. To accelerate the simulation of adverse weather effects, previously several models were introduced for camera \cite{Halder_2019_ICCV, sakaridisfog} and lidar data \cite{rasshofer_influences_2011, hasirlioglu_novel_2020,kilic_lidar_2021,hahner_fog_2021}. 

Simulated adverse weather effects are fog ~\cite{rasshofer_influences_2011, hahner_fog_2021,sakaridisfog}, rain \cite{hasirlioglu_novel_2020,kilic_lidar_2021, Halder_2019_ICCV} and snow \cite{snow_hahner_2022}. 
Compared to these weather effects, object dependent environmental influences such as spray caused by individual road participants appear more frequently \cite{walz_benchmark_2021}. For spray, only wet roads are required. Wet roads appear up to 100 times a year in comparison to heavy rainfall (5 times a year), dense fog (12 times a year) and blizzards (13 times a year) \cite{walz_benchmark_2021}. Despite of having low to no rainfall, spray causes devastating degeneration of lidar sensors and subsequent deep learning models used for detecting other road participants \cite{walz_benchmark_2021,shearman_trials_1998}. 
In particular, spray plumes are often falsely classified as, e.g., vehicles, which would effectively render autonomous driving under spray conditions impossible.
However, the simulation model can help to improve the robustness of these detectors by generating sufficient amounts of training data corrupted by spray.
Simulated data has the advantage that it includes interesting edge-cases recorded only in, e.g., clear weather conditions.
Furthermore, it avoids the expensive process of data collection and labeling.

Hence, this publication deals with the simulation of spray effects for lidar sensors which can be used as training data for object detection algorithms.
In summary, this publication

\begin{itemize}
    \item presents a novel data set taken on a test track with method ical variation of influencing parameters of spray,
    \item introduces a probabilistic phenomenological modeling approach for spray in lidar data, focusing on clustering of detection points in the spray plume, without the need of fluid dynamics simulation or physics engines,
    \item validates the spray simulation method by utilizing it as a data augmentation method for two deep learning approaches,
    \item shows that the simulation approach increases object detection under spray significantly for two state-of-the art object detectors on a diverse real world spray test set consisting of 1463 frames. 
\end{itemize}
The \href{https://github.com/openMSL/reflection_based_lidar_object_model/tree/master/src/model/strategies/lidar-environmental-effects-strategy}{\underline{simulation model}}\footnote{https://github.com/openMSL/reflection-based-lidar-object-model/tree/master/src/model/strategies/lidar-environmental-effects-strategy}, the \href{https://gitlab.com/tuda-fzd/data-augmentation/lidar-spray-augmentation}{\underline{augmentation}}\footnote{https://gitlab.com/tuda-fzd/data-augmentation/lidar-spray-augmentation} as well as the \href{https://www.fzd-datasets.de/spray/}{\underline{recorded data set}}\footnote{https://www.fzd-datasets.de/spray/} are all open-source and publicly available.

\section{Related Work}

A first systematic study on the influence of road spray on radar and lidar was published by Shearman~et~al.~\cite{shearman_trials_1998} in 1998.
Measurements were taken on a watered test track with cars and trucks at different speeds.
Attenuation as well as clutter due to spray were observed in lidar data while radar was only marginally affected.
Walz~et~al.~\cite{walz_benchmark_2021} developed a portable spray machine that can be attached to different vehicles in order to produce spray in a reproducible and controlled manner.
By testing overtaking maneuvers, the authors also observed a strong influence on lidar in form of detections from the spray plume.
Another effect is partial blockage due to water on the sensors surface, when it is inside the spray plume of a leading or cutting vehicle.
In addition, Walz~et~al. tested an object detection algorithm on the lidar data and found that detections from spray cause false positive objects to be detected posing a safety risk for automated vehicles.

To replicate spray in simulation, previously two simulation methods were introduced in \cite{vargas_rivero_data_2021,vargas_rivero_effect_2021,shih_reconstruction_2022}. 
The first method by Rivero~et~al.~\cite{vargas_rivero_effect_2021},~\cite{vargas_rivero_data_2021} uses a renderer with a physics engine in Blender.
The spray simulation is done by generating particles which do not represent the actual water droplets but rather depend on the number of detections observed in real spray measurements.
These particles are ejected from the tires backwards with a physically calculated velocity while considering wind and drag forces.
Turbulence simulation is also considered with a Perlin texture, but it is not further elaborated on and it becomes not apparent in the data.
In contrast, our evaluations of turbulence caused by the moving vehicle reveal its high influence on the distribution and velocities of the spray plume.
Furthermore, our data shows that turbulence leads to the formation of distinct clusters in the spray plume, which move in the direction of the moving vehicle.
This has not been described in literature, yet.
That might be due to the fact that Vargas Rivero~et~al. used a low resolution \textit{Ibeo Scala}, where the clustering effect is not as visible as in higher resolution lidars.
For validation, accumulated point clouds and histograms are used there, which also hide the inhomogeneity of detections from spray in single time steps.
While the results look very promising for the used low resolution lidar, the adaptation to other sensor configurations is not viable.

Shih~et~al.~\cite{shih_reconstruction_2022} introduced a data-driven approach to synthesize spray in lidar data.
A precomputed list of particle movements is matched to real spray data.
Variations of particles are generated and rendered with a ray casting algorithm simulating the lidar beams.
The pipeline is applied to augment point clouds recorded in clear weather conditions and used to generate training data for a machine learning object detection algorithm.
The data used for reconstruction is taken from the Waymo dataset~\cite{sun_scalability_2020}, which for the most part does not contain isolated spray but is mixed with rain.
For validation, a point-to-point comparison of the synthetic point cloud with the real data is used.

In contradiction to previous work, our method investigates the introduced disturbance patterns of spray plumes in detail through two specifically captured data sets for high resolution automotive laser scanning systems.
The disturbance is analyzed and modeled by clustering of the point cloud in a first data set with isolated spray effects and the overall model validation is performed using object detection algorithms on a second data set.

\section{Method}

\begin{figure}[!t]
 	\includegraphics[width=\columnwidth]{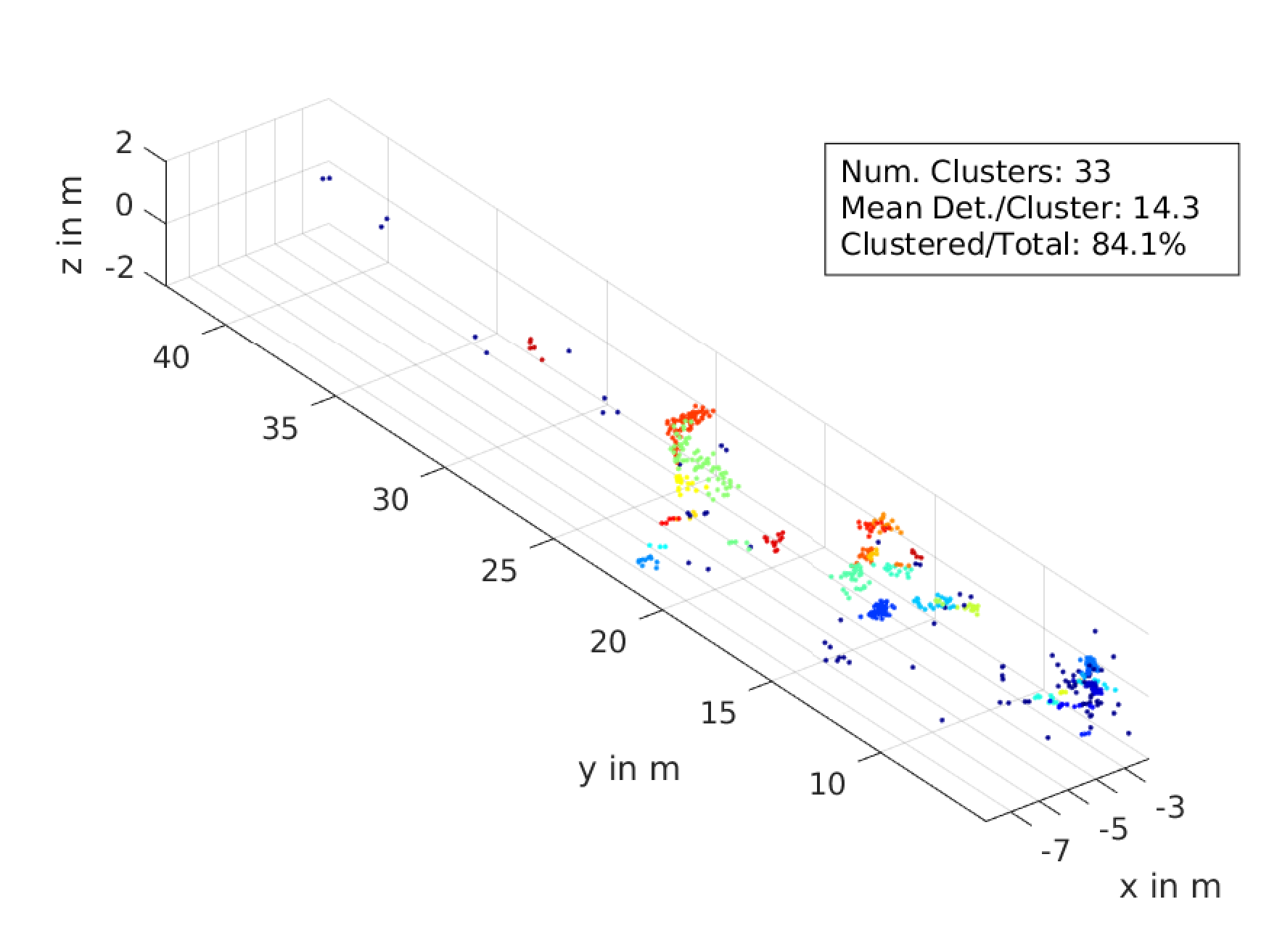}
 	\caption{Measured point cloud of a spray plume generated with a VW Crafter at $50\,\si{\metre}$ with a velocity of $\SI{100}{\kilo\metre/\hour}$. A DBSCAN clustering algorithm is applied to the point cloud and the colors in the plot represent individual clusters. The detections are in the sensor coordinate system with the y-axis pointing towards the forward direction of the ego vehicle.}
     \label{fig:clustering_2}
 \end{figure}

\begin{figure}[!t]
	\centering
	\scalebox{1.0}{%
		\input{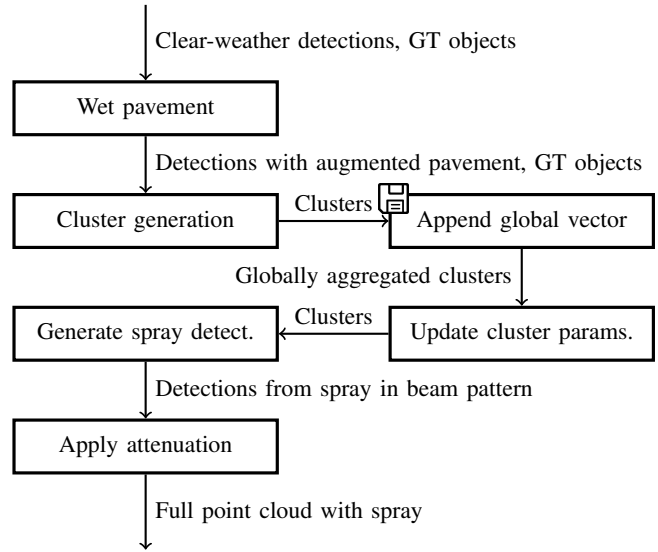}
	}
	\caption{Structure of the spray simulation module. Detections and ground truth objects are the input from which clusters are generated for every time step. The clusters are appended to a global cluster vector and updated every time step. At beam interactions with the clusters, detections are generated and finally an attenuation is applied to all beams going through the spray plume. GT = Ground Truth}
    \label{fig:model_structure_2}
\end{figure}

The method introduced in this work enables to transform clear weather recordings or simulated data to point clouds with spray replicating real world disturbances in lidar measurements. Thereby, the model is based on a phenomenological description to simplify the generation of spray plumes, instead of simulating the aerodynamics, particle distributions and laser ray interactions through computationally expensive Monte Carlo simulations. 
Dominating factors determining the properties of a whirled up spray plume are the vehicle velocity, vehicle class, the dwell time of independent particles, distance of the leading vehicle and lastly the water level height. \figref{fig:clustering_2} qualitatively presents the introduced artifacts from a spray plume. The build up of aerodynamic vertices with higher particle densities cause the formation of individual spray clusters. Based on such observations, the simulation method depicted in \figref{fig:model_structure_2} is split into individual building blocks. 

The input to the model are  sensor detections from a clear-weather scene as well as information about ground truth objects, e.g. vehicle dimensions, positions and velocities.
The first step is the simulation of the wet pavement, as road spray does not occur without water on the pavement. This first step leads to significant amounts of lost ground points due to specular reflection of laser beams on watered surfaces. To simulate this loss, the wet ground model from \cite{snow_hahner_2022} is used.
Secondly, the individual clusters are generated containing a cluster center, time since generation and position. The initialization of clusters is restricted to the vehicle trajectory and no clusters are spawned randomly in the scene. The registered time allows the clusters to dissolve over time. 

The state of every cluster kept in the global vector is updated in every time step the spray simulation is applied. The update rule is discussed in greater detail in section \ref{par:clusterUpdate}.
These first two steps of the algorithm keep track of the disturbances introduced into the scene. Thirdly, the interaction with the  sensor is modeled, where a lidar beam hits a cluster and measures either a false detection from the cluster with a certain likelihood or causes attenuation of the laser beam intensity.
To conform with the scanner beam pattern, a ray casting algorithm is used, which takes into account all clusters along every ray.

After registering false detections and the attenuation of each ray interacting with the generated clusters, an augmented point cloud with additional disturbances from whirled up spray is outputted.
Each of those steps depends on a set of parameters which will be introduced in the following section beginning with an in-depth description on the calibration data set captured on a proving ground for individual parameters.

Note, current automotive lidar sensors are capable of generating multiple detections per beam.
However, state-of-the-art object detection algorithms rely on a single strongest return, see e.g. \cite{openpcdet}. Thus, only the strongest return is considered in the presented method.

\subsection{Spray Experiments and Calibration Data Set}

\begin{figure*}[ht]
     \centering
     \begin{subfigure}[b]{0.48\textwidth}
         \centering
         \includegraphics[width=\textwidth]{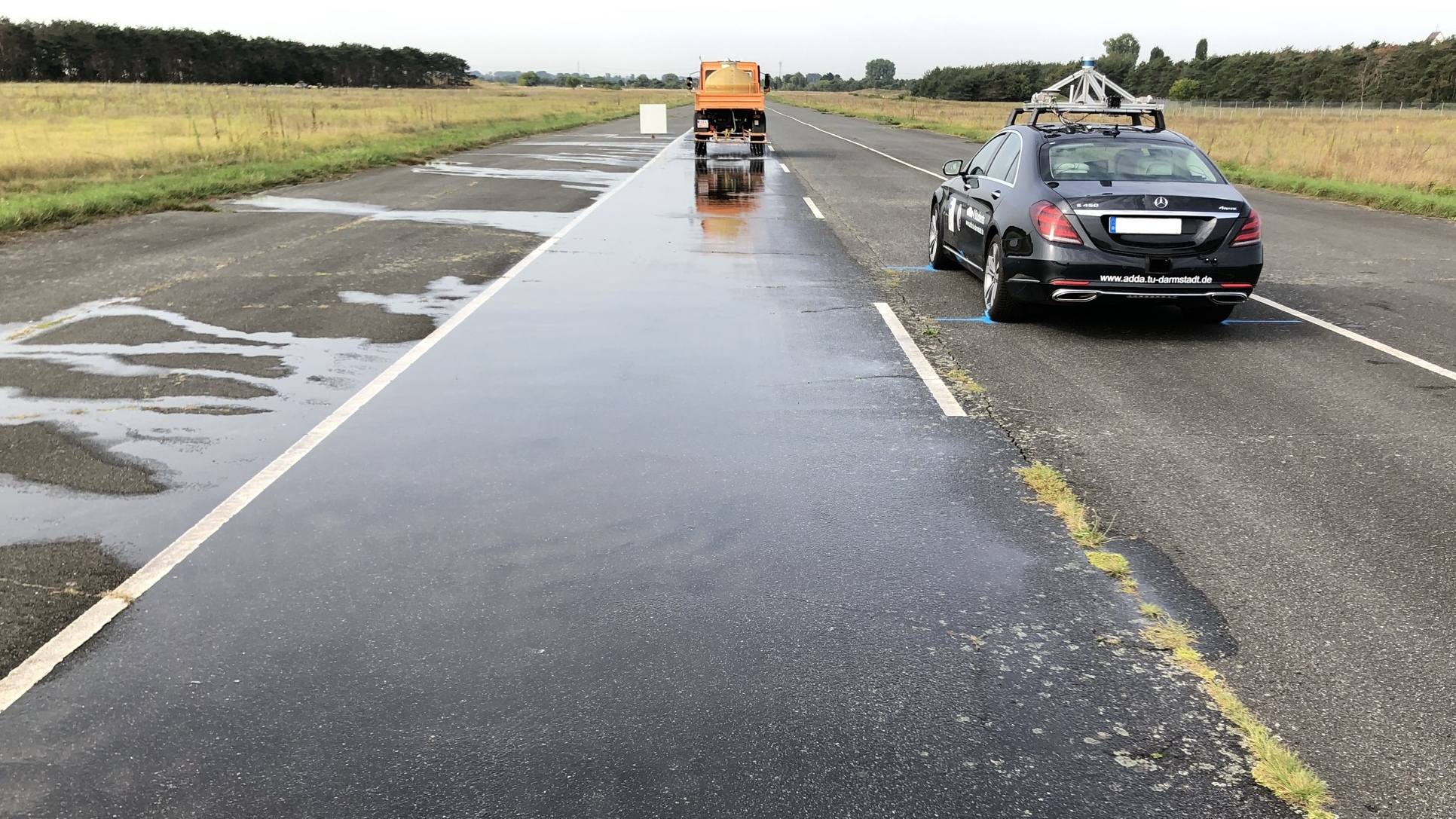}
         \caption{Pavement preparations with a watering system mounted on a Unimog. The measurement vehicle Mercedes S-Class is parked on the right and the lidar target on the left.}
         \label{fig:unimog2}
     \end{subfigure}
     \hfill
     \begin{subfigure}[b]{0.48\textwidth}
         \centering
         \includegraphics[width=\textwidth]{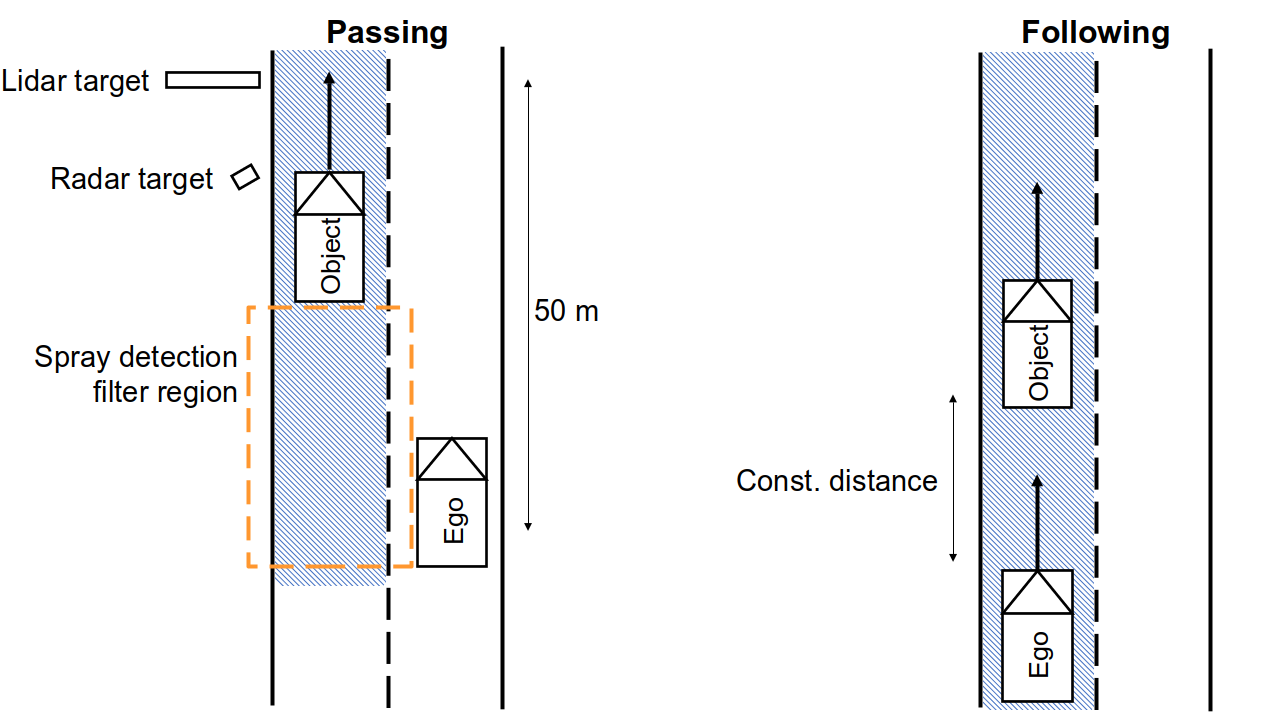}
         \caption{Scenario sketches for passing (left) and following (right)}
         \label{fig:scenarios2}
     \end{subfigure}
        \caption{Experimental setup on August-Euler-Airfield test track in Darmstadt, Germany}
        \label{fig:experiment2}
\end{figure*}

To calibrate the model, extensive experiments are conducted to capture sufficient amount of reference data. The tests are designed to capture introduced spray effects in lidar data as isolated as possible. Therefore, tests are conducted on an isolated test track with a watering vehicle preparing the test track equally for each iteration. The scene is watered by a Unimog, as shown in \figref{fig:unimog2}. Next to the Unimog, the test vehicle, equipped with a \textit{Velodyne VLP32} on top of its roof, can be seen.
The pavement is watered by pumping water from a tank through a distribution system at the rear of the Unimog while maintaining constant pump and vehicle speed.
Object vehicles are equipped with an RTK-based GNSS-IMU system, recording pose and velocities as reference during the experiments. In addition, defined object positions are marked and a reference lidar and radar target are placed next to the watered road in $50\,\si{\metre}$ distance.
As target vehicles, two different types are used and driven over the watered lane following the scenarios depicted in \figref{fig:scenarios2}.

In the first scenario, the ego vehicle is stationary on the right lane and an object vehicle is passing on the left lane.
The direct line of sight from the sensor to the reference target is going through the spray plume.
In the second scenario, the ego vehicle is following the object vehicle with constant speed. 
Following Pilkington~et~al.~\cite{meyer_splash_1990}, the subsequent parameters are identified as most influential on the spray plume: object velocity, object class, water film height, as well as distance are therefore varied during the measuring campaign.
Similar to Pilkington~et~al, spray plumes start to appear at a vehicle speed of $50\,\si{\kilo\metre/\hour}$, setting this as minimal speed for our analysis. 
To cover most world wide speed limits in real world traffic, a maximum speed for data collection is set to $130\,\si{\kilo\metre/\hour}$.
To keep the number of experiments in a realistic time frame, the velocity interval is set to $10\,\si{\kilo\metre/\hour}$.
For varying the object class, the experiments are carried out with a VW Golf representing a compact car and a VW Crafter as a large van.

Note that throughout the test, it was noticed that the surface of the road shows imperfections as cracks, small hills and valleys which are evened out by the water and lead to an inhomogeneous distribution of the water film. This was also reported by Vargas Rivero~et~al.~\cite{vargas_rivero_effect_2021}.
Therefore, precisely mapping out the water film height is not feasible. Instead it was taken care to cover three classes of wetness levels: moist, partially water covered and fully water covered. 
To ensure equal conditions throughout all measurements, the water was reapplied at the beginning of each measurement step. 

After watering, the object vehicle is driven 6 times over the pavement, reducing the water film height with every iteration.
The pavement watering is subsequently categorized in the three classes described previously by a human operator.

\subsection{Calibration}

Based on the conducted experiments, the individual parts of the spray simulation model are calibrated.
Following \figref{fig:model_structure_2}, a parameterization needs to be found that describes
\begin{itemize}
    \item the generation and distribution of clusters
    \item the update of cluster locations and their movement over time
    \item the generation of spray detections in the lidar
    \item the attenuation of other objects behind spray plumes.
\end{itemize}
To calibrate the model in \figref{fig:model_structure_2}, the experiments are reproduced in the simulation tool IPG CarMaker.
Data is transferred between simulation tool and model using the \gls{osi}.
In CarMaker, reflections are generated by a super-sampled ray tracing, which are then utilized by a reflection-based lidar model \cite{rosenberger_reflection_2022}.
In the model, the reflections are sorted into beams and detections are extracted with a peak detection algorithm.
Alongside the reflections, the model also receives ground truth information about all moving and stationary objects within the vicinity of the sensor.
The reflection-based model was developed in a highly modular framework~\cite{linnhoff_highly_2021}, enabling extensions in form of contained strategies that can be added into the modeling pipeline.

\PAR{Cluster Generation.} As indicated by \figref{fig:clustering_2}, spray is organized into clusters.
To quantify this effect, a DBSCAN~\cite{ester_density-based_1996} algorithm is applied to the detections.
Over \SI{84}{\percent} of the detections in \figref{fig:clustering_2} are sorted into clusters which is also confirmed by the other measurements of the data set.

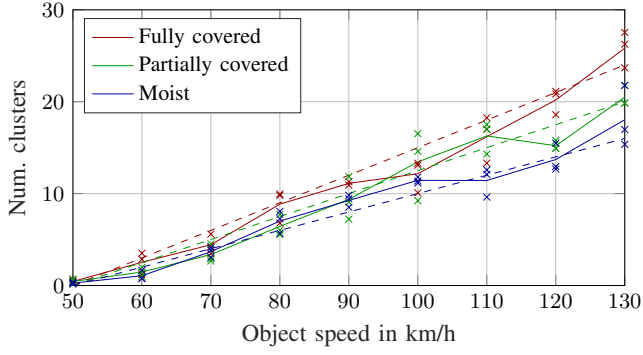
\begin{figure}[ht]
    \resizebox{\columnwidth}{!} {
%
%
\begin{tikzpicture}[font=\small]

\begin{axis}[%
width=8cm,
height=4cm,
at={(0.758in,0.481in)},
scale only axis,
xmin=50,
xmax=130,
xlabel style={font=\color{white!15!black}},
xlabel={Object speed in km/h},
ymin=0,
ymax=30,
ylabel style={font=\color{white!15!black}},
ylabel={Num. clusters},
axis background/.style={fill=white},
xmajorgrids,
ymajorgrids,
legend style={at={(0.02,0.97)}, anchor=north west, legend cell align=left, align=left, draw=white!15!black}
]
\addplot [color=black!40!red]
  table[row sep=crcr]{%
50	0.441696113074205\\
60	2.52747252747253\\
70	4.43599493029151\\
80	8.85375494071146\\
90	11.1111111111111\\
100	12.1602288984263\\
110	16.1958146487294\\
120	20.1877934272301\\
130	25.8292282430213\\
};
\addlegendentry{Fully covered}

\addplot [color=black!40!green]
  table[row sep=crcr]{%
50	0.441696113074205\\
60	1.46520146520147\\
70	3.37135614702155\\
80	6.42951251646904\\
90	9.38271604938272\\
100	13.447782546495\\
110	16.2780269058296\\
120	15.2112676056338\\
130	20.4926108374384\\
};
\addlegendentry{Partially covered}

\addplot [color=black!40!blue]
  table[row sep=crcr]{%
50	0.265017667844523\\
60	1.06227106227106\\
70	3.68187579214195\\
80	7.00922266139657\\
90	9.25925925925926\\
100	11.444921316166\\
110	11.4275037369208\\
120	13.7089201877934\\
130	18.0377668308703\\
};
\addlegendentry{Moist}

\addplot [color=black!40!red, only marks, mark=x, mark options={solid, black!40!red}]
  table[row sep=crcr]{%
50	0.265017667844523\\
50	0.441696113074205\\
50	0.618374558303887\\
60	3.51648351648352\\
60	1.20879120879121\\
60	2.85714285714286\\
70	4.25855513307985\\
70	3.46007604562738\\
80	6.79841897233202\\
80	9.96047430830039\\
80	9.80237154150198\\
90	11.2962962962963\\
90	10.9259259259259\\
100	10.0858369098712\\
100	13.3047210300429\\
100	13.0901287553648\\
110	17.0179372197309\\
110	18.2511210762332\\
110	13.3183856502242\\
120	20.8450704225352\\
120	18.5915492957747\\
120	21.1267605633803\\
130	23.6945812807882\\
130	26.256157635468\\
130	27.5369458128079\\
70	5.5893536121673\\
};

\addplot [color=black!40!green, only marks, mark=x, mark options={solid, black!40!green}]
  table[row sep=crcr]{%
50	0.176678445229682\\
50	0.441696113074205\\
50	0.706713780918728\\
60	1.97802197802198\\
60	1.42857142857143\\
70	2.92775665399239\\
70	2.6615969581749\\
70	4.52471482889734\\
80	6.00790513833992\\
80	7.74703557312253\\
80	5.53359683794466\\
90	11.8518518518519\\
90	9.07407407407407\\
90	7.22222222222222\\
100	16.5236051502146\\
100	9.2274678111588\\
100	14.5922746781116\\
110	17.0179372197309\\
110	17.5112107623318\\
110	14.304932735426\\
120	14.9295774647887\\
120	15.7746478873239\\
120	14.9295774647887\\
130	21.7733990147783\\
130	19.8522167487685\\
130	19.8522167487685\\
60	0.989010989010989\\
};

\addplot [color=black!40!blue, only marks, mark=x, mark options={solid, black!40!blue}]
  table[row sep=crcr]{%
50	0.176678445229682\\
50	0.265017667844523\\
50	0.353356890459364\\
60	1.64835164835165\\
60	0.769230769230769\\
70	3.85931558935361\\
70	3.06083650190114\\
70	4.1254752851711\\
80	8.06324110671937\\
80	5.69169960474308\\
80	7.27272727272727\\
90	9.44444444444444\\
90	8.51851851851852\\
90	9.81481481481481\\
100	11.1587982832618\\
100	11.8025751072961\\
100	11.3733905579399\\
110	12.085201793722\\
110	12.5784753363229\\
110	9.61883408071749\\
120	12.9577464788732\\
120	15.4929577464789\\
120	12.6760563380282\\
130	16.9704433497537\\
130	15.3694581280788\\
130	21.7733990147783\\
60	0.769230769230769\\
};

\addplot [color=black!40!red, dashed]
  table[row sep=crcr]{%
50	0.0\\
130	24.0\\
};

\addplot [color=black!40!green, dashed]
  table[row sep=crcr]{%
50	0.0\\
130	20.0\\
};

\addplot [color=black!40!blue, dashed]
  table[row sep=crcr]{%
50	0.0\\
130	16.0\\
};

\end{axis}

\end{tikzpicture}%
    }
    \caption{Measured number of clusters generated by the VW Crafter during one scan. Values from single measurements are marked as x and mean values per velocity are denoted as lines. The dashed lines represent linear approximations of the mean used for model calibration.}
    \label{fig:num_clusters}
\end{figure}

For the simulation model, the number of clusters that are generated in each simulation time step are calculated as a function of the object speed, class, and water film height.
This is analyzed by evaluating the detections from spray in the data set, where the vehicle is close to the sensor.
An object distance of \SI{25}{\metre} from the sensor is chosen, since here distinct and stable clusters are observed with high resolution in the sensor.
Using the DBSCAN algorithm, clusters are identified and counted.
The DBSCAN algorithm is applied to the point cloud in spherical coordinates to account for the change in Cartesian resolution over distance.
As small clusters are observed in the recorded data, three points is defined as the minimum number of points that can form a cluster.
In the central elevation field of view, the Velodyne VLP32 has an elevation angle discretization of $0.333^{\circ}$.
To ensure possible clustering in vertical direction, the search neighborhood of DBSCAN is set to $0.7^{\circ}$ to cover at least two scan lines of the sensor.
The number of clusters generated within one simulation time step is determined by first counting the number of clusters in a spray filter region behind the vehicle.
This total number is then divided by the length ratio of the spray region to the distance the object traveled during one scan of the sensor.
As the frame rate of the real sensor is equal to the rate of simulation time steps, this results in a mean number of clusters generated during one time step for every object speed of both object classes.
The result for the VW Crafter is plotted in \figref{fig:num_clusters}.
The dashed lines are linear approximations of the mean values used as the expected values during cluster generation.

\begin{figure}[ht]
    \resizebox{\columnwidth}{!} {
        \input{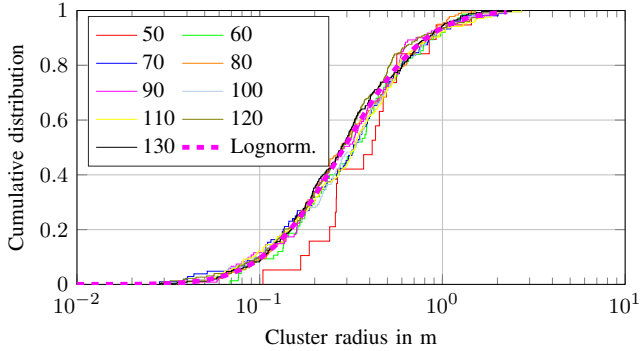}
    }
    \caption{Cluster radius distribution in the measurements for different speeds in km/h denoted as colors. The dashed distribution is a lognormal distribution derived as a mean from fitted distributions to the measurements of all velocities. The parameters of the distribution are $\mu = -1.2$ and $\sigma = 0.8$ as an example.}
    \label{fig:clusters_radius_dist}
\end{figure}

Next, the cluster size is calibrated.
Since the clusters are approximated by spheres, the radii $r_\text{cl}$ are set to the maximum distance of detections to the respective cluster center.
The results of the cluster distribution from the VW Crafter measurements with fully covered pavement at different object speeds are shown in \figref{fig:clusters_radius_dist}.
A lognormal distribution is chosen as the best fit by Matlab's \textit{fitdist} function.
The distribution parameters are evaluated for every object speed and the mean values are taken for the overall distribution for the model. 

To simulate clusters, spheres are positioned with a uniform distribution in the region behind every moving object according to the empirical values for the number of clusters and the distribution of cluster radii.
A uniform distribution was chosen as a first estimate, as no other form of distribution was visible.
Nevertheless, this will be evaluated in future work.
With the independent uniform placement of clusters with different radii, overlapping will occur in random cases.
On one hand, overlapping will reduce the calibrated number of clusters.
On the other hand, it increases inhomogeneity even further by creating other shapes than spheres.
The final model evaluation will show that the theoretical reduction in clusters will not be significant in the end result, so overlapping is not prohibited in the model.

Next, the dynamic aspects of clusters are defined, as clusters do not stay in one position.
Similar to Shearman~et~al.~\cite{shearman_trials_1998}, the measurements show that clusters first move in the direction of the moving object, then decelerate due to air resistance.
This is possibly caused by the dead wake directly behind the moving object.
Hence, during generation of a cluster, the velocity of the moving object is set as an initial cluster velocity.
In the presence of wind, the clusters move in the wind direction.
For now, wind direction and speed are presumed the same for all clusters, although wind gusts could have a varying effect on different clusters.

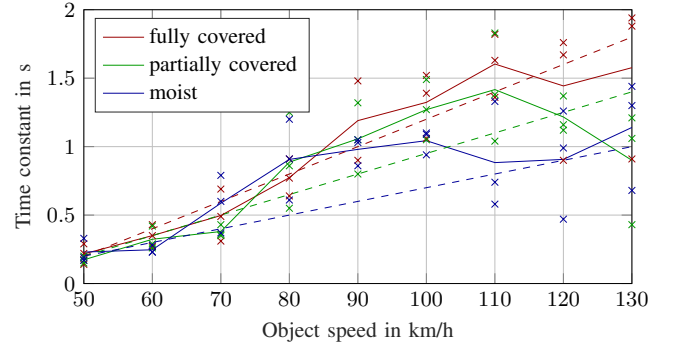
\begin{figure}[t]
    \resizebox{\columnwidth}{!} {
%
%
\begin{tikzpicture}

\begin{axis}[%
width=8cm,
height=4cm,
at={(0.758in,0.481in)},
scale only axis,
xmin=50,
xmax=130,
xlabel style={font=\color{white!15!black}\small},
xlabel={Object speed in km/h},
ymin=0,
ymax=2,
ylabel style={font=\color{white!15!black}\small},
ylabel={Time constant in s},
axis background/.style={fill=white},
xmajorgrids,
ymajorgrids,
legend style={at={(0.02,0.97)}, anchor=north west, legend cell align=left, align=left, draw=white!15!black, font=\small}
]
]
\addplot [color=black!40!red]
  table[row sep=crcr]{%
50	0.216666666666667\\
60	0.35\\
70	0.496666666666667\\
80	0.773333333333333\\
90	1.19\\
100	1.32333333333333\\
110	1.60333333333333\\
120	1.44333333333333\\
130	1.57666666666667\\
};
\addlegendentry{fully covered}

\addplot [color=black!40!green]
  table[row sep=crcr]{%
50	0.173333333333333\\
60	0.323333333333333\\
70	0.38\\
80	0.886666666666667\\
90	1.05666666666667\\
100	1.27\\
110	1.41666666666667\\
120	1.21666666666667\\
130	0.9\\
};
\addlegendentry{partially covered}

\addplot [color=black!40!blue]
  table[row sep=crcr]{%
50	0.23\\
60	0.246666666666667\\
70	0.586666666666667\\
80	0.906666666666667\\
90	0.98\\
100	1.04333333333333\\
110	0.883333333333333\\
120	0.906666666666667\\
130	1.14\\
};
\addlegendentry{moist}

\addplot [color=black!40!red, only marks, mark=x, mark options={solid, black!40!red}]
  table[row sep=crcr]{%
50	0.14\\
60	0.35\\
70	0.49\\
80	0.91\\
90	1.48\\
100	1.06\\
110	1.82\\
120	0.9\\
130	1.94\\
50	0.29\\
60	0.27\\
70	0.31\\
80	0.64\\
100	1.39\\
110	1.63\\
120	1.67\\
130	0.91\\
50	0.22\\
60	0.43\\
70	0.69\\
80	0.77\\
90	0.9\\
100	1.52\\
110	1.36\\
120	1.76\\
130	1.88\\
};

\addplot [color=black!40!red, dashed]
  table[row sep=crcr]{%
50	0.2\\
130	1.8\\
};

\addplot [color=black!40!green, only marks, mark=x, mark options={solid, black!40!green}]
  table[row sep=crcr]{%
50	0.17\\
60	0.42\\
70	0.36\\
80	1.25\\
90	0.8\\
100	1.49\\
110	1.38\\
120	1.16\\
130	1.21\\
50	0.15\\
60	0.29\\
70	0.35\\
80	0.86\\
90	1.32\\
100	1.27\\
110	1.83\\
120	1.37\\
130	0.43\\
50	0.2\\
60	0.26\\
70	0.43\\
80	0.55\\
90	1.05\\
100	1.05\\
110	1.04\\
120	1.12\\
130	1.06\\
};

\addplot [color=black!40!green, dashed]
  table[row sep=crcr]{%
50	0.2\\
130	1.4\\
};

\addplot [color=black!40!blue, only marks, mark=x, mark options={solid, black!40!blue}]
  table[row sep=crcr]{%
50	0.17\\
60	0.28\\
70	0.37\\
80	0.61\\
90	0.86\\
100	0.94\\
110	1.33\\
120	1.26\\
130	1.3\\
50	0.19\\
60	0.23\\
70	0.6\\
80	0.91\\
90	1.05\\
100	1.1\\
110	0.74\\
120	0.99\\
130	1.44\\
50	0.33\\
60	0.23\\
70	0.79\\
80	1.2\\
90	1.03\\
100	1.09\\
110	0.58\\
120	0.47\\
130	0.68\\
};

\addplot [color=black!40!blue, dashed]
  table[row sep=crcr]{%
50	0.2\\
130	1.0\\
};

\end{axis}

\end{tikzpicture}%
    }
    \caption{Time constants taken at 64 \% of numbers of detection from the spray plume with respect to the maximum of the measurement. Values from single measurements are marked as x and mean values per velocity are denoted as lines. The dashed lines represent linear approximations of the mean used for model calibration.}
    \label{fig:time_constants}
\end{figure}

Furthermore, clusters dissolve in the air over time.
The time it takes for a cluster to dissolve, such that it does not generate detections any more, varies from cluster to cluster.
Clusters do not just disappear after a certain amount of time, but they dissolve gradually.
This behavior is modeled by changing the detection probability of the clusters exponentially over time with a probability factor based on an exponential impulse response
\begin{equation}
    \text{P}_\text{t}(t) = e^{-\frac{t}{T}},
    \label{eq:time_factor}
\end{equation}
where the time constant $T$ is defined as the time required to reduce the detection count to 36\,\% of the maximum.
To determine $T$, the numbers of detections from spray inside a constant volume filter are taken over time for different object speeds and watering levels.
From \figref{fig:time_constants}, it becomes apparent that the time constants rise with respect to object speed, which is approximated by linear functions for the different watering levels.
The watering levels are parameterized in the simulation with \SI{0.5}{\milli\metre}, \SI{0.75}{\milli\metre} and \SI{1.0}{\milli\metre} for moist, partially covered, and fully covered, respectively.
These are approximations based on previous exploratory measurements with a pavement wetness sensor.

\PAR{Update Cluster Parameters.} \label{par:clusterUpdate} The next step is to update the cluster parameters from previous time steps in the global cluster vector.
An age counter is increased by the simulation time step size $\Delta t$.
The cluster velocity $\vec{v}_\text{cl,k}$ is calculated for the current simulation time step $k$.
As mentioned, the cluster velocity is influenced by wind, drag and an initial velocity. 
Wind forces and drag forces are defined as
\begin{equation}
    \vec{F}_\text{D} = \dfrac{1}{2} \rho_\text{air} c_\text{D} A \vec{v}^2_\text{rel}
\end{equation}
with the air density $\rho_\text{air}$, drag coefficient $c_\text{D}$, cross section of the droplets $A$ and the relative wind velocity $\vec{v}_\text{rel}$.
The relative wind velocity consists of the overall wind velocity $\vec{v}_\text{w}$ and the cluster velocity $\vec{v}_\text{cl}$.
This results in the dynamic equilibrium with the mass $m$ of a droplet and its acceleration $a$ yields
\begin{equation}
    m \cdot \vec{a} = \dfrac{1}{2} \rho_\text{air} c_\text{D} A (\vec{v}_\text{w}+\vec{v}_\text{cl})^2.
\end{equation}
Integrating the equation yields the velocity
\begin{equation}
    \vec{v}_\text{cl} = \dfrac{\rho_\text{air} c_\text{D} A}{2m} \int (\vec{v}_\text{w}+\vec{v}_\text{cl})^2 dt + \vec{v}_\text{0}
\end{equation}
with the initial velocity of the cluster $\vec{v}_\text{0}$.
A step-wise discrete integration yields the velocity of the cluster in time step $k$
\begin{equation}
    \vec{v}_\text{cl,k} = \text{C} \cdot (\vec{v}_\text{w}+\vec{v}_\text{cl,k-1})^2 \cdot \Delta t + \vec{v}_\text{cl,k-1}
\end{equation}
with the constant factor \text{C} containing the unknown values of the cluster particle mass, droplet cross section, drag coefficient and the air density.
The constant is  determined experimentally by comparing the theoretical calculation with the cluster movement in real measurements.
The vehicle velocity and wind velocity are measured, leaving the constant as the only unknown.
Optimizing the constant to fit the movement of the clusters in measurements with different vehicle velocities and wind velocities yields a factor of $\text{C} = \SI{-0.15}{\per\metre}$.
It should be noted that the factor depends on the step size $\Delta t$ and has to be re-calibrated for sensors with different scanning frequencies.
The here applied step size according to the sensors frequency is $\Delta t = \SI{100}{\milli\second}$.
Using the calculated cluster velocity, the current position is updated with
\begin{equation}
    \vec{p}_\text{cl,k} = \vec{p}_\text{cl,k-1} + \Delta t \cdot \vec{v}_\text{cl,k}.
\end{equation}

\begin{figure}[t]
    \resizebox{\columnwidth}{!} {
        \input{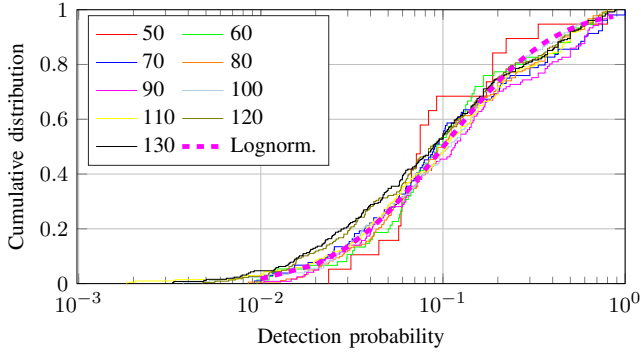}
    }
    \caption{Measured distribution of detection probabilities per beam hitting a spherical cluster representation taken for all speeds in km/h denoted as colors. The dashed distribution is a lognormal distribution derived as a mean from fitted distributions to the measurements of all velocities. The parameters of the distribution are $\mu= -2.3$ and $\sigma= 1.09$.}
    \label{fig:detection_probability_dist}
\end{figure}

\PAR{Generate Spray Detections.} Subsequently, spray detections using the updated clusters are generated.
With a ray casting algorithm, the intersections between lidar beams and spherical clusters are calculated.
A beam generates only a detection in a certain cluster with a detection probability given as
\begin{equation}
    \text{P}_{\text{det}} = \dfrac{n}{n_{\text{max}}(r_\text{cl},\vec{p}_\text{cl})},
\end{equation}
with the number of detections $n$ in a given cluster and  the number of beams $n_{\text{max}}(r_\text{cl},\vec{p}_\text{cl})$ intersecting the spherical representation of the cluster.
The maximum number of possible detections in a spherical cluster representation depends on the cluster radius and the cluster position with respect to the beam pattern of the sensor.
The measured detection probability distributions for all clusters in the spray plume with a vehicle distance of \SI{25}{\metre} are shown in \figref{fig:detection_probability_dist}.
The variations of detection probabilities between clusters are fitted to a lognormal distribution.
The distribution parameters for the overall distribution are the mean values of the parameters of every respective object speed.
As the cluster dissolves, the detection probability decreases by the time factor $T$ introduced in equation \eqref{eq:time_factor}.
If a detection is generated according to the probability, the range is set to a normal distribution around the middle of the cluster and added to a vector, as one beam can generate multiple detections when hitting multiple cluster.
Over 80\,\% of spray detections from the VLP32 have an intensity value of $0$ and 16\,\% are $1$.
Therefore, the final intensity value of the simulated spray detections are set to $0$.
However, to allow for variation and an intensity comparison between multiple detections in one beam, a normal distributed value with a mean value of $0.5$ and a standard deviation of $0.05$ is calculated during modeling.

\PAR{Attenuation.} Since a beam intersecting with a cluster generates only a detection with a certain detection probability, a beam can pass multiple clusters, before generating a detection.
As water in the air not only reflects light back to the sensor but also scatters the light, an attenuation occurs.

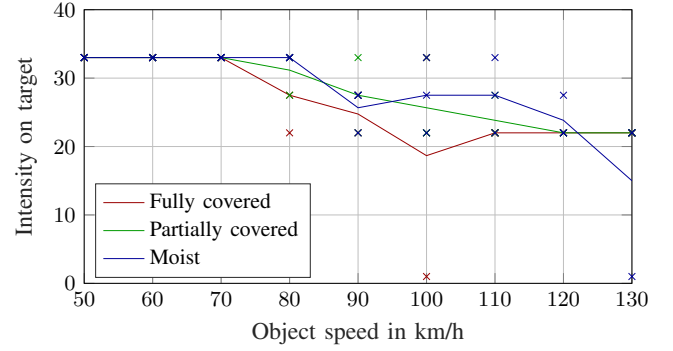
\begin{figure}[t]
    \resizebox{\columnwidth}{!} {
%
%
\begin{tikzpicture}

\begin{axis}[%
width=8cm,
height=4cm,
at={(0.758in,0.481in)},
scale only axis,
xmin=50,
xmax=130,
xlabel style={font=\color{white!15!black}},
xlabel={Object speed in km/h},
ymin=0,
ymax=40,
ylabel style={font=\color{white!15!black}},
ylabel={Intensity on target},
axis background/.style={fill=white},
xmajorgrids,
ymajorgrids,
legend style={at={(0.02,0.02)}, anchor=south west, legend cell align=left, align=left, draw=white!15!black}
]
\addplot [color=black!40!red]
  table[row sep=crcr]{%
50	33\\
60	33\\
70	33\\
80	27.5\\
90	24.75\\
100	18.6666666666667\\
110	22\\
120	22\\
130	22\\
};
\addlegendentry{Fully covered}

\addplot [color=black!40!green]
  table[row sep=crcr]{%
50	33\\
60	33\\
70	33\\
80	31.1666666666667\\
90	27.5\\
100	25.6666666666667\\
110	23.8333333333333\\
120	22\\
130	22\\
};
\addlegendentry{Partially covered}

\addplot [color=black!40!blue]
  table[row sep=crcr]{%
50	33\\
60	33\\
70	33\\
80	33\\
90	25.6666666666667\\
100	27.5\\
110	27.5\\
120	23.8333333333333\\
130	15\\
};
\addlegendentry{Moist}

\addplot [color=black!40!red, only marks, mark=x, mark options={solid, black!40!red}]
  table[row sep=crcr]{%
50	33\\
50	33\\
50	33\\
60	33\\
60	33\\
60	33\\
70	33\\
70	33\\
80	33\\
80	22\\
80	27.5\\
90	22\\
90	27.5\\
100	1\\
100	33\\
100	22\\
110	22\\
110	22\\
110	22\\
120	22\\
120	22\\
120	22\\
130	22\\
130	22\\
130	22\\
70	33\\
};

\addplot [color=black!40!green, only marks, mark=x, mark options={solid, black!40!green}]
  table[row sep=crcr]{%
50	33\\
50	33\\
50	33\\
60	33\\
60	33\\
70	33\\
70	33\\
70	33\\
80	33\\
80	33\\
80	27.5\\
90	22\\
90	33\\
90	27.5\\
100	22\\
100	33\\
100	22\\
110	22\\
110	22\\
110	27.5\\
120	22\\
120	22\\
120	22\\
130	22\\
130	22\\
130	22\\
60	33\\
};

\addplot [color=black!40!blue, only marks, mark=x, mark options={solid, black!40!blue}]
  table[row sep=crcr]{%
50	33\\
50	33\\
50	33\\
60	33\\
60	33\\
70	33\\
70	33\\
70	33\\
80	33\\
80	33\\
80	33\\
90	27.5\\
90	27.5\\
90	22\\
100	22\\
100	33\\
100	27.5\\
110	27.5\\
110	22\\
110	33\\
120	22\\
120	22\\
120	27.5\\
130	22\\
130	1\\
130	22\\
60	33\\
};

\end{axis}

\end{tikzpicture}%
    }
    \caption{Measured mean intensities of detections on the lidar target. The intensities are recorded when the rear of the passing VW Crafter is next to the target. Values from single scans are marked as x and mean values from 3 measurements for every speed are denoted as lines.}
    \label{fig:intensities_on_target}
\end{figure}
The total distance a beam traveled through multiple clusters is calculated and an attenuation is applied.
A first analysis of the decrease of detection distances in spray for bigger spray clouds in terms of cluster count as well as a decrease in detection intensity on the lidar target yields a mean extinction coefficient of $\SI{0.02}{\per\metre}$.
This value was found iteratively as an optimization to match the decrease in intensity on the simulated target to the measurements shown in \figref{fig:intensities_on_target}.
According to calculations by Goodin et al.\,\cite{goodin_predicting_2019} based on measurements with a Velodyne VLP16, this corresponds to the attenuation in moderate rainfall with an intensity of around $\SI{3}{\milli\metre/\hour}$.
If a detection existed in a beam before the spray simulation, the attenuation due to the spray is also applied to its intensity.
In the end, the detection intensities of all detections from a beam are compared and the strongest is added to the overall point cloud.

The demonstrated calibration on the example of the Crafter is performed the same way for the Golf vehicle.
However, the parameters of cluster radius, detection probability and intensity stay within the variations within one vehicle class.
The only differences between the two vehicles are the number of clusters and the dissolve time constant.
These are set separately for the object classes.

\subsection{Application in Object Detection}

Following recent work \cite{sakaridisfog,Halder_2019_ICCV,snow_hahner_2022,hahner_fog_2021}, a viable method to verify the realism of simulation methods is its application as data augmentation method for downstream object detection tasks.
To this end, the simulation model is used to generate synthetic training data that is corrupted by spray.
It is shown that the additional training data alone is enough to significantly improve the performance of lidar-based object detectors in real-world spray scenarios.

To validate the made approximations several experiments are run including comparisons to related augmentation methods for fog \cite{hahner_fog_2021} and rain \cite{kilic_lidar_2021}. 
Additionally, the augmentation performance is compared to a removal of cluttered points through \cite{charron_-noising_2018}. 
Moreover, two different object detectors PV-RCNN \cite{shi_pvrcnn} and Voxel-RCNN \cite{deng_voxel} are investigated. 
Both are implemented within the OpenPCDet codebase \cite{openpcdet}.
These detectors are chosen since they are state-of-the art and provide the largest benefit in \cite{snow_hahner_2022}, where the effects of snow on lidar-based object detectors are investigated.

\PAR{Training.} The training is performed on the \gls{stf} data set. 
As spray occurs only in correspondence with vehicles, great emphasis is put on vehicles and other object classes are omitted.
The original training and validation splits for clear conditions provided with \gls{stf} data set are used. 
Hence, 3469 clear weather point clouds are available for training.
However, the spray augmentation is only applied to samples where the ego-velocity exceeds \SI{60}{\kilo\metre/\hour}.
Furthermore, only highway and suburban scenarios are considered for augmentation.
Analog to \cite{snow_hahner_2022}, all detectors are trained from scratch with the default configurations for each method.

\PAR{Data Augmentation.} In order to utilize the simulation model for augmentation, the input requirements of the model need to be met.
The model requires knowledge of the trajectories of the ego vehicle and all surrounding vehicles that cause spray.
This information is not readily available in the \gls{stf} data set.
Thus, the velocity from surrounding vehicles is approximated by the velocity information in the corresponding radar detections.
The trajectories are subsequently computed by backwards integration over a fixed time horizon of \SI{5}{\second}.
This assumption has proven to be accurate enough for scenarios where spray typically occurs, e.g., on the highway.
Since the simulation model is a function of the water level, the water level is sampled from an exponential distribution from the interval \SI{0.1}-\SI{1.2}{\milli\metre} during training.

\PAR{Baseline Methods.} As baseline methods, the spray augmentation is compared to the clear weather baseline.
Furthermore, related adverse weather augmentations models are investigated, for rain \cite{kilic_lidar_2021} and for fog \cite{hahner_fog_2021}.
Both methods are parameterized as described in \cite{snow_hahner_2022}.
Moreover, the de-noising algorithm DROR \cite{charron_-noising_2018} is compared. 
DROR is applied before the point cloud is given as input to the object detector removing clutter from spray plumes.

\PAR{Metrics.} Finally, the effectiveness of the spray augmentation is evaluated on a dedicated spray test set.
The collection of this test set is detailed in \ref{ch:spray_testset}.
For evaluation, the KITTI evaluation framework \cite{kitti_2012} with the standard Intersection-over-Union (IoU) threshold is used.
A suitable evaluation metric must consider the inherent conflict between precision and recall. 
Commonly, average precision (AP) is used  for this in object detection tasks. 
A conclusive overview of evaluation metrics can be found in \cite{survey_ap_2020}.
Analog to \cite{snow_hahner_2022}, AP at 40 recall position is reported.
Since spray is highly dependent on distance to the ego vehicle, AP is reported depending on different ranges as done in \cite{gated3d_2021, snow_hahner_2022}.
Since the test set is labeled  up to \SI{50}{\metre}, 0-\SI{50}{\metre} can be considered the main evaluation range.
Furthermore, the performance on ranges 0-\SI{30}{\metre} and 30-\SI{50}{\metre} is reported.
Additionally, the object detectors are evaluated on the clear weather test set defined in the \gls{stf} data set.
All trainings and subsequent evaluations are run three times and averaged to reduce statistical fluctuations.

\subsection{Spray Test Set for Object Detection}
\label{ch:spray_testset}

A fair evaluation of the spray  for measuring the performance increase is only possible on collected real-world scenes that include spray.
Such data sets are not readily available.
Thus, as a first step, a test set dedicated to spray is created.
This is a twofold process including data collection and labeling.
In total, 1463 frames are collected and labeled.

\PAR{Data Collection.} For collecting real-world spray scenes, an additional measurement campaign on a closed highway test track is conducted.
The same acquisition vehicle as for the \gls{stf} data set is used.
Since this data set is used during training, differences regarding sensor makes and mounting positions between the training and newly created test set are eliminated.
Measurement runs consist of following other vehicles on wet roads.
As shown previously, spray is heavily affected by vehicle size, velocity, water level, and distance to ego vehicle.
Hence, these parameters are varied during the measurement runs.
An overview of the collected data is provided in \tabref{tab:sprayset}.
In summary, three different vehicles in different scenarios are being followed - namely, a Mercedes-Benz G-Class being the largest, a Mercedes-Benz CLA-Class being a mid-sized, and a Smart being the smallest vehicle. 
The speed of the ego vehicle is kept constant during a measurement run.
Target speeds are 80, 100, and \SI{120}{\kilo\metre/\hour}.
Furthermore, it is distinguished between two road conditions.
Measurements with wet road conditions are conducted immediately after rain.
Additionally, the test track provides the possibility to artificially water the road with high water contents close to aquaplaning conditions.
Thus, the resulting water film is considerably larger compared to wet road conditions.
To capture a significant number of spray effects, the followed vehicles are close to the ego vehicle in distances from 0-50 m.
\begin{table}
	\centering
	\caption{Overview of the collected spray data set. Car describes the type of car being followed. Ego speed is the average speed plus standard deviation of the ego vehicle. Samples are collected on a wet road after rain and an artificially watered road. Frames describes the specific number of frames being used in the overall test set.}
	\label{tab:sprayset}
	\begin{tabular}{l r r r}
		\textbf{Vehicle} & \textbf{Ego speed} in km/h & \textbf{Road Condition} & \textbf{Frames} \\
		\hline \hline \noalign{\vskip 1mm}
		G-Class & $79.9 \pm2.8$ & wet & 324 \\
		Smart & $118.3  \pm{6.0}$ & wet & 248 \\
		CLA & $97.4 \pm{2.7}$ & wet & 170 \\
		Smart & $97.6 \pm{4.1}$ & watered & 149\\
		CLA & $121.0 \pm{3.7}$ & watered & 319\\
		CLA + 2x G-Class & $96.2 \pm{1.7}$ & watered & 253
	\end{tabular}
\end{table}

\PAR{Auto-Labeling Pipeline.} For labeling, an auto-label pipeline tailored specifically towards data affected by spray is implemented.
As indicated by, e.g., \figref{fig:qual_results}, the outputs from the PV-RCNN detector trained without augmentation show that an lidar-based object detector alone is not sufficient for generating these labels.
The auto-generated labels would suffer from a large number of false positives, as well as imprecise location and dimension information.
To mitigate these effects, the auto-label pipeline relies on a tracker and additional radar measurements.
This allows to track the vehicle over an extended period of time and validate the detections with the radar.
Here, the radar acts as a ground truth sensor as previous work has shown that it is not substantially affected by spray and therefore ideally suited to safeguard the object detections \cite{walz_benchmark_2021}.
This way erroneous false detections from spray are suppressed since they are not consistent in consecutive frames and not detected by a radar sensor. 

\begin{figure}[t]
	\centering
	\scalebox{1.0}{%
		\begin{tikzpicture}[scale=1.0]

\tikzset{
    block/.style={
           rectangle,
           align=center,
           draw=black, very thick,
           minimum height=2em,
           minimum width=2.6cm,
           inner sep=4pt,
           text centered,
           font=\small,
           node distance=1.5cm
           },
    storage/.style={
           rectangle,
           rounded corners,
           draw=black, very thick,
           minimum height=2em,
           minimum width=3.5cm,
           inner sep=2pt,
           text centered,
           font=\small,
           fill={rgb, 255:red, 200; green, 200; blue, 200 }
           },
    block_legend/.style={
           rectangle,
           draw=black, very thick,
           minimum height=1.5em,
           minimum width=1.5cm,
           inner sep=4pt,
           text centered,
           font=\small,
           node distance=0.65cm
           },
    storage_legend/.style={
           rectangle,
           rounded corners,
           draw=black, very thick,
           minimum height=1.5em,
           minimum width=1.5cm,
           inner sep=2pt,
           text centered,
           font=\small,
           fill={rgb, 255:red, 200; green, 200; blue, 200 }
           },
    label/.style={
           font=\small
           },
}

\pgfmathsetmacro{\DIST}{4.5cm}

\node[label, align=center] (Input) {Point cloud};
\node[label, right of=Input, node distance=\DIST] (Radar) {Radar data};
\node[block, below of=Input, node distance=1cm] (ObjDet) {Object detection};
\node[block, below of=ObjDet] (DataAssoc) {Data association};
\node[block, right of=DataAssoc, node distance=\DIST] (RadarAssoc) {Data association};
\node[block, below of=DataAssoc] (Kalman) {Kalman filter};
\node[block, below of=RadarAssoc] (Memory) {Memory};
\node[block, below of=Memory, align=center] (Sup) {Suppression};
\node[label, below of=Sup] (Output) {Ground truth objects};

\pgfmathsetmacro{\ARRSHIFT}{0.15cm}

\draw[->,thick] (Input.south) -- (ObjDet.north);
\draw[->,thick] (Radar.south) -- (RadarAssoc.north);
\draw[->,thick] (ObjDet.south) -- (DataAssoc.north) node[label, midway, left, align=center] {3D\\objects};

\draw[->,thick] ([xshift=-\ARRSHIFT] DataAssoc.south) -- ([xshift=-\ARRSHIFT] Kalman.north) node[label, midway, left] {Update};
\draw[->,thick] ([xshift=\ARRSHIFT] Kalman.north) -- ([xshift=\ARRSHIFT] DataAssoc.south) node[label, midway, right] {Predict};

\draw[->, thick, bend left] (DataAssoc.east) to node[label, pos=0.35, above=0.15cm, align=center] {New,\\delete}(Memory.north west);

\draw[->,thick] ([xshift=-\ARRSHIFT] RadarAssoc.south) -- ([xshift=-\ARRSHIFT] Memory.north) node[label, midway, left] {Update};
\draw[->,thick] ([xshift=\ARRSHIFT] Memory.north) -- ([xshift=\ARRSHIFT] RadarAssoc.south) node[label, midway, right, align=center] {BeV\\objects};

\draw[->,thick] (Memory.south) -- (Sup.north) node[label, midway, right] {Trajectories};
\draw[->,thick] (Memory.west) -- (Kalman.east) node[label, midway, above] {3D objects};
\draw[->,thick] (Sup.south) -- (Output.north);

\node (fit) [draw, dashed, inner sep=2.0mm, label=below:Rectangle Title, fit=(ObjDet) (Kalman) (Memory) (Kalman)] {};
\node [label, dashed, yshift=-0.0cm, align=center] at (fit.north) {Repeat for\\each frame}; 


\end{tikzpicture}
	}
	\caption{Auto-labeling pipeline for labeling the spray test set. The original AB3DMOT tracker is extended with radar data that is associated to trajectories outputted by the Kalman filter. Trajectories which are not assigned a sufficient number of radar detections are suppressed.}
    \label{fig:label_gen}
\end{figure}
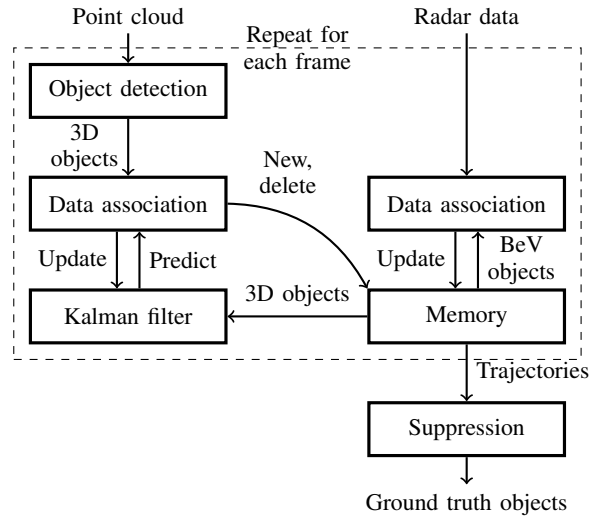

To this end, the multi-object tracker presented in \cite{weng_3d_2020} is adapted.
This tracker, subsequently referred to as AB3DMOT, combines 3D object detections from a lidar-based object detector with a Kalman filter. 
A general overview of AB3DMOT is part of \figref{fig:label_gen}. 
The 3D objects predicted by the object detector are associated to the prediction from a Kalman filter.
For the prediction, the Kalman filter relies on a memory that saves the trajectories of the previously tracked vehicles. 
Successfully associated objects are then used to update the Kalman filter and subsequently the corresponding trajectory.
Objects that cannot be associated are used to initialize a new vehicle trajectory.
Trajectories that have not been updated for a certain time are deleted from the memory. 
The straightforward design allows for low computation times while achieving state-of-the-art tracking results.
However, on the collected spray data, the performance of AB3DMOT is insufficient due to the large number of false positives and disappearing detection only recognizable in the radar sensor. 

\PAR{Extending AB3DMOT with Radar.} Thus, the original AB3DMOT tracker is extended by considering detections from a radar.
As shown in \figref{fig:label_gen}, radar detections are first associated to a tracked vehicle. 
Analogous to the data association problem in AB3DMOT, the Hungarian algorithm is used for finding the best match between radar detections and tracked vehicles. 
As an association metric, the smallest distance of the radar detection to any edge of the vehicle is used.
To simplify this computation, only the birds-eye view of the vehicle is considered.
When a radar detection is successfully associated to a tracked vehicle, a counter is incremented.
The counter is later used to suppress trajectories without a sufficient number of assigned radar detections.
This means that velocity and location information of the radar detection are not fused directly in the Kalman filter.
In experiments, the fusion of this information did not prove beneficial for the quality of the labels.
This is likely related to the fact that using this information requires, e.g., knowledge of the object's heading.
Since this is an output of the object detector, the introduced noise likely outweighs the benefit.

The process of detecting, associating, and updating is repeated for each frame.
However, in contrast to online tracking problems, the tracked objects can be adjusted after the fact.
This is used to provide temporally more consistent labels.
To be more precise, after the tracker has seen each frame, it is looped through all trajectories saved in the memory.
As mentioned previously, every object belonging to a trajectory without a sufficient number of radar detections is suppressed.
This is indicated at the bottom of \figref{fig:label_gen}.
For the considered scenarios, this suppression effectively eliminates all false positives consistently. 
Furthermore, the same vehicle dimensions are assigned to vehicles that are tracked over multiple frames.
To compute the vehicles dimensions, the dimension from the detections with the 20 highest confidence scores are used.  
Finally, after the label generation, each scenario is assessed by a human annotator and it is found that no manual corrections had to be performed. 

\begin{figure}[t]
     \centering
     \begin{subfigure}[b]{0.49\columnwidth}
         \centering
         \includegraphics[width=\columnwidth]{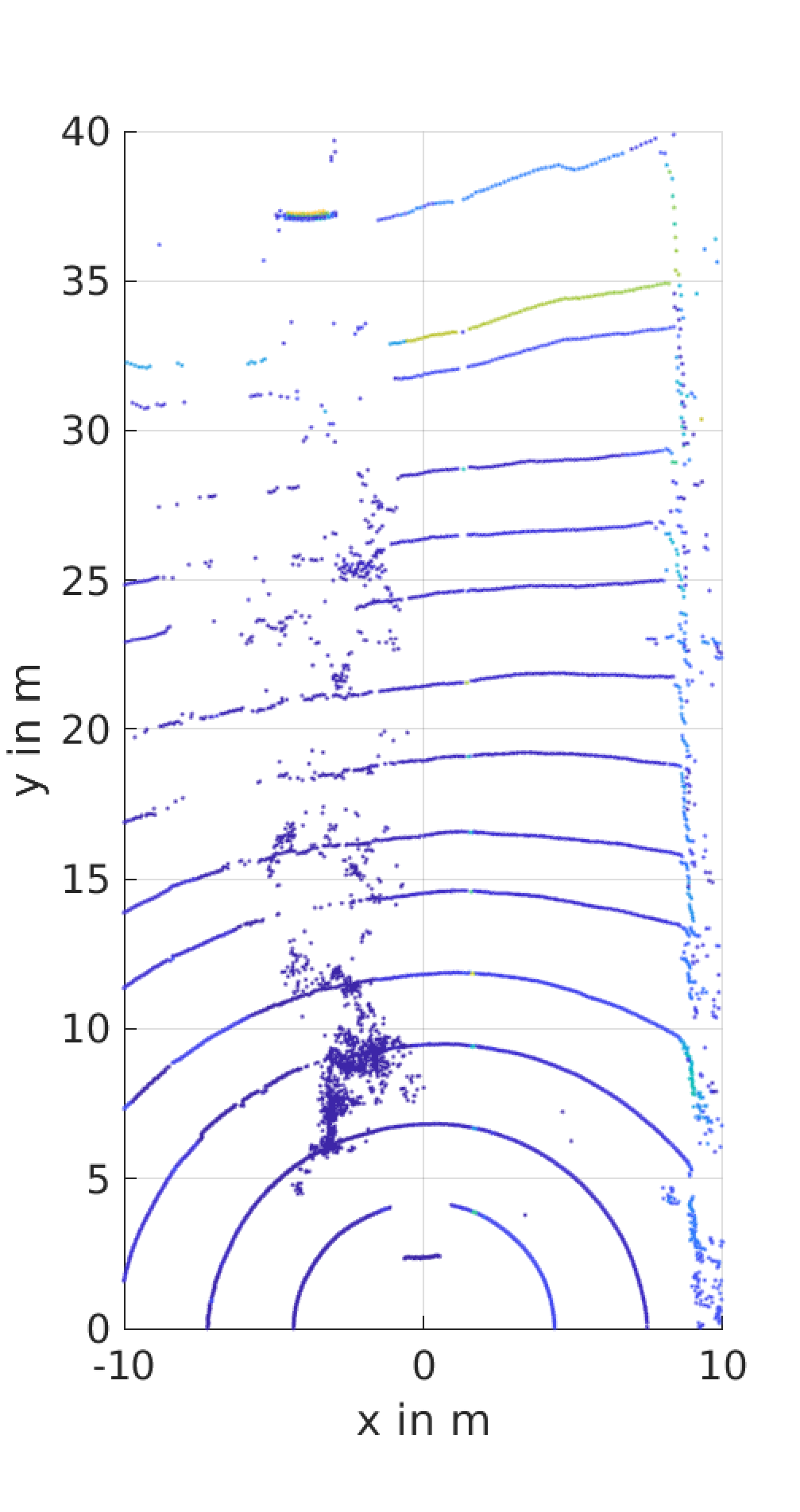}
         \caption{Measured point cloud}
         \label{fig:eval_pcl_real}
     \end{subfigure}
     \hfill
     \begin{subfigure}[b]{0.49\columnwidth}
         \centering
         \includegraphics[width=\columnwidth]{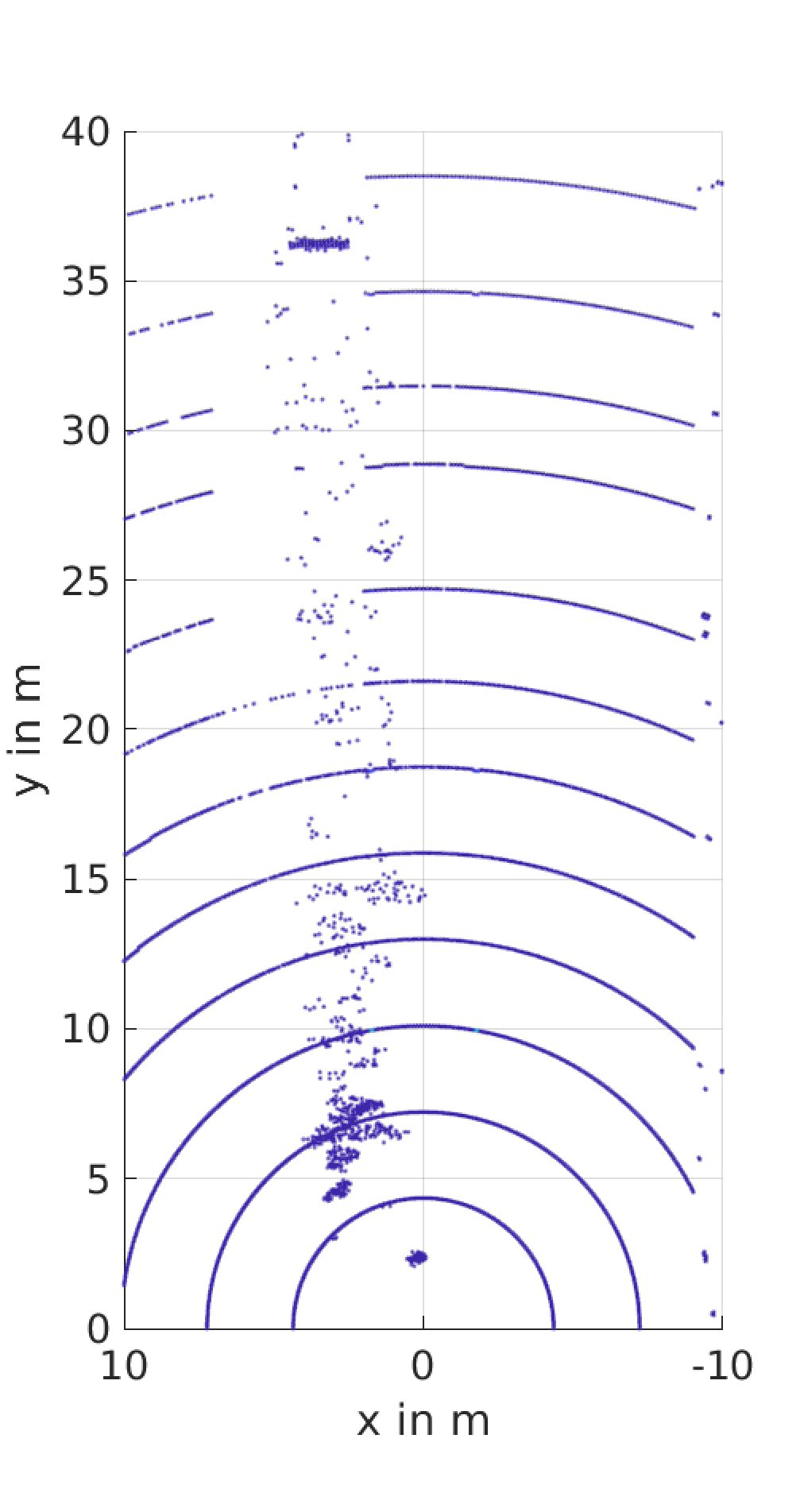}
         \caption{Simulated point cloud}
         \label{fig:eval_pcl_sim}
     \end{subfigure}
        \caption{Point clouds of a single scan from a real measurement and the simulation of a VW Crafter passing by the stationary ego vehicle. The VW Crafter is driving at a speed of $\SI{100}{\kilo\metre/\hour}$ and is located in this scan at around  $\SI{40}{\metre}$. The origin of these plots is the origin of the sensor.}
        \label{fig:eval_pcl_compare} 
\end{figure}

\section{Results}

In this section, first the results from the simulation model are compared to the calibration data set. Then, result from training object detection algorithm with augmented data are presented.

\subsection{Simulation Results}

To evaluate the model, object distances and values, which were not utilized for the model calibration, are used on the same data set.
This is done to evaluate and verify the model and its calibration.
First, a qualitative comparison of the simulated point cloud with a measured point cloud is shown in \figref{fig:eval_pcl_compare}. Here, both a real world scan \figref{fig:eval_pcl_real} and simulation \figref{fig:eval_pcl_sim} of a passing VW Crafter with a vehicle speed of $\SI{100}{\kilo\metre/\hour}$ at a distance of $\SI{40}{\metre}$ is depicted.
Qualitatively speaking, the inhomogeneity of both point clouds stands out with detections forming clusters in different shapes, sizes and densities.
Furthermore, the number of detections decreases over the distance.

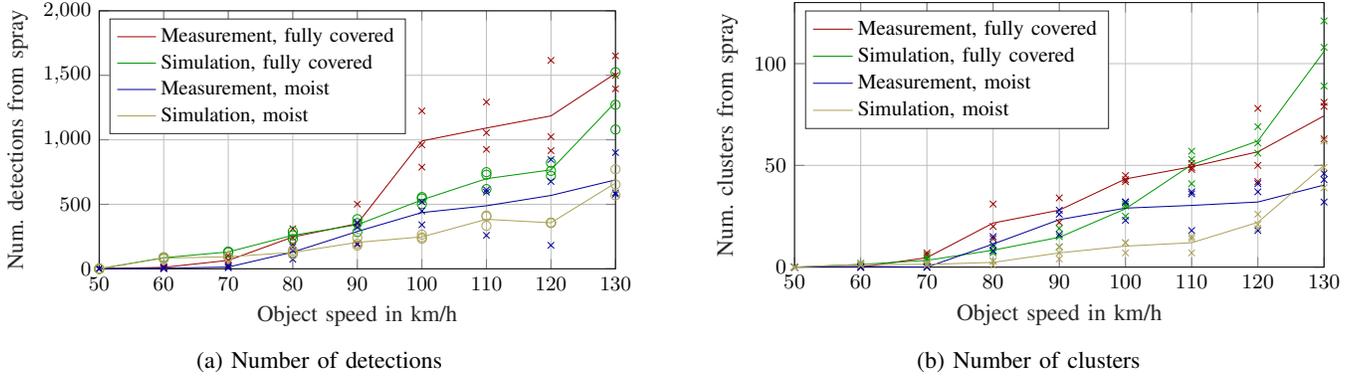
\begin{figure*}[ht]
     \centering
     \begin{subfigure}[b]{0.48\textwidth}
         \centering
         \resizebox{\columnwidth}{!} {
%
%
\begin{tikzpicture}

\begin{axis}[%
width=8cm,
height=4cm,
at={(0.758in,0.481in)},
scale only axis,
xmin=50,
xmax=130,
xlabel style={font=\color{white!15!black}},
xlabel={Object speed in km/h},
ymin=0,
ymax=2000,
ylabel style={font=\color{white!15!black}},
ylabel={Num. detections from spray},
axis background/.style={fill=white},
xmajorgrids,
ymajorgrids,
legend style={at={(0.02,0.97)}, anchor=north west, legend cell align=left, align=left, draw=white!15!black}
]
\addplot [color=black!40!red]
  table[row sep=crcr]{%
50	1.33333333333333\\
60	11\\
70	65\\
80	246.333333333333\\
90	349.666666666667\\
100	991\\
110	1091.33333333333\\
120	1185.33333333333\\
130	1514.66666666667\\
};
\addlegendentry{Measurement, fully covered}

\addplot [color=black!40!green]
  table[row sep=crcr]{%
50	0\\
60	85.6666666666667\\
70	130.666666666667\\
80	259.666666666667\\
90	341.333333333333\\
100	534.333333333333\\
110	698.666666666667\\
120	767\\
130	1292\\
};
\addlegendentry{Simulation, fully covered}

\addplot [color=black!40!blue]
  table[row sep=crcr]{%
50	0\\
60	4.33333333333333\\
70	13.3333333333333\\
80	129.333333333333\\
90	289.333333333333\\
100	436\\
110	488.333333333333\\
120	568.666666666667\\
130	689.333333333333\\
};
\addlegendentry{Measurement, moist}

\addplot [color=black!40!yellow]
  table[row sep=crcr]{%
50	0\\
60	82.6666666666667\\
70	93.3333333333333\\
80	126.666666666667\\
90	204.666666666667\\
100	247.333333333333\\
110	383.666666666667\\
120	356.666666666667\\
130	665\\
};
\addlegendentry{Simulation, moist}

\addplot [color=black!40!red, only marks, mark=x, mark options={solid, black!40!red}]
  table[row sep=crcr]{%
50	0\\
50	2\\
50	2\\
60	16\\
60	4\\
60	13\\
70	66\\
70	28\\
80	188\\
80	311\\
80	240\\
90	501\\
90	198\\
90	350\\
100	788\\
100	961\\
100	1224\\
110	926\\
110	1293\\
110	1055\\
120	916\\
120	1616\\
120	1024\\
130	1395\\
130	1651\\
130	1498\\
70	101\\
};

\addplot [color=black!40!green, only marks, mark=o, mark options={solid, black!40!green}]
  table[row sep=crcr]{%
50	0\\
50	0\\
50	0\\
60	86\\
60	89\\
60	82\\
70	132\\
70	128\\
70	132\\
80	260\\
80	230\\
80	289\\
90	285\\
90	354\\
90	385\\
100	556\\
100	502\\
100	545\\
110	749\\
110	618\\
110	729\\
120	720\\
120	822\\
120	759\\
130	1080\\
130	1272\\
130	1524\\
};

\addplot [color=black!40!blue, only marks, mark=x, mark options={solid, black!40!blue}]
  table[row sep=crcr]{%
50	0\\
50	0\\
50	0\\
60	3\\
60	1\\
70	10\\
70	9\\
70	21\\
80	143\\
80	74\\
80	171\\
90	359\\
90	188\\
90	321\\
100	519\\
100	341\\
100	448\\
110	594\\
110	612\\
110	259\\
120	847\\
120	677\\
120	182\\
130	901\\
130	589\\
130	578\\
60	9\\
};

\addplot [color=black!40!yellow, only marks, mark=o, mark options={solid, black!40!yellow}]
  table[row sep=crcr]{%
50	0\\
50	0\\
50	0\\
60	77\\
60	83\\
60	88\\
70	90\\
70	94\\
70	96\\
80	127\\
80	112\\
80	141\\
90	240\\
90	194\\
90	180\\
100	265\\
100	233\\
100	244\\
110	413\\
110	405\\
110	333\\
120	358\\
120	358\\
120	354\\
130	654\\
130	569\\
130	772\\
};

\end{axis}

\end{tikzpicture}%
        }
         \caption{Number of detections}
         \label{fig:eval_num_detections}
     \end{subfigure}
     \hfill
     \begin{subfigure}[b]{0.48\textwidth}
         \centering
         \resizebox{\columnwidth}{!} {
%
%
\begin{tikzpicture}

\begin{axis}[%
width=8cm,
height=4cm,
at={(0.758in,0.481in)},
scale only axis,
xmin=50,
xmax=130,
xlabel style={font=\color{white!15!black}},
xlabel={Object speed in km/h},
ymin=0,
ymax=130,
ylabel style={font=\color{white!15!black}},
ylabel={Num. clusters from spray},
axis background/.style={fill=white},
xmajorgrids,
ymajorgrids,
legend style={at={(0.02,0.97)}, anchor=north west, legend cell align=left, align=left, draw=white!15!black}
]
\addplot [color=black!40!red]
  table[row sep=crcr]{%
50	0\\
60	0\\
70	4.66666666666667\\
80	21.6666666666667\\
90	28\\
100	43.3333333333333\\
110	49.3333333333333\\
120	56.6666666666667\\
130	74.3333333333333\\
};
\addlegendentry{Measurement, fully covered}

\addplot [color=black!40!green]
  table[row sep=crcr]{%
50	0\\
60	1.33333333333333\\
70	3.33333333333333\\
80	8.33333333333333\\
90	14.6666666666667\\
100	28.6666666666667\\
110	50.3333333333333\\
120	62\\
130	106\\
};
\addlegendentry{Simulation, fully covered}

\addplot [color=black!40!blue]
  table[row sep=crcr]{%
50	0\\
60	0.333333333333333\\
70	0\\
80	11.3333333333333\\
90	23.3333333333333\\
100	29\\
110	30.3333333333333\\
120	32\\
130	40.3333333333333\\
};
\addlegendentry{Measurement, moist}

\addplot [color=black!40!yellow]
  table[row sep=crcr]{%
50	0\\
60	1.33333333333333\\
70	1.33333333333333\\
80	2.33333333333333\\
90	7\\
100	10.3333333333333\\
110	12\\
120	22\\
130	50\\
};
\addlegendentry{Simulation, moist}

\addplot [color=black!40!red, only marks, mark=x, mark options={solid, black!40!red}]
  table[row sep=crcr]{%
50	0\\
50	0\\
50	0\\
60	0\\
60	0\\
60	0\\
70	6\\
70	1\\
80	20\\
80	31\\
80	14\\
90	34\\
90	22\\
100	43\\
100	45\\
100	42\\
110	48\\
110	49\\
110	51\\
120	42\\
120	78\\
120	50\\
130	63\\
130	81\\
130	79\\
70	7\\
};

\addplot [color=black!40!green, only marks, mark=x, mark options={solid, black!40!green}]
  table[row sep=crcr]{%
50	0\\
50	0\\
50	0\\
60	2\\
60	1\\
60	1\\
70	1\\
70	4\\
70	5\\
80	8\\
80	7\\
80	10\\
90	10\\
90	15\\
90	19\\
100	30\\
100	25\\
100	31\\
110	57\\
110	41\\
110	53\\
120	56\\
120	69\\
120	61\\
130	89\\
130	108\\
130	121\\
};

\addplot [color=black!40!blue, only marks, mark=x, mark options={solid, black!40!blue}]
  table[row sep=crcr]{%
50	0\\
50	0\\
50	0\\
60	0\\
60	0\\
70	0\\
70	0\\
70	0\\
80	11\\
80	8\\
80	15\\
90	28\\
90	16\\
90	26\\
100	32\\
100	23\\
100	32\\
110	36\\
110	37\\
110	18\\
120	37\\
120	41\\
120	18\\
130	46\\
130	43\\
130	32\\
60	1\\
};

\addplot [color=black!40!yellow, only marks, mark=x, mark options={solid, black!40!yellow}]
  table[row sep=crcr]{%
50	0\\
50	0\\
50	0\\
60	1\\
60	1\\
60	2\\
70	1\\
70	1\\
70	2\\
80	3\\
80	1\\
80	3\\
90	4\\
90	7\\
90	10\\
100	12\\
100	7\\
100	12\\
110	14\\
110	15\\
110	7\\
120	26\\
120	21\\
120	19\\
130	39\\
130	49\\
130	62\\
};

\end{axis}

\end{tikzpicture}%
        }
         \caption{Number of clusters}
         \label{fig:eval_num_clusters}
     \end{subfigure}
        \caption{Comparison of simulated and measured numbers of detections and numbers of clusters in the spray plume from a Crafter at $\SI{50}{\metre}$ distance}
        \label{fig:eval_detections_clusters}
\end{figure*}

For a quantitative analysis, the total number of detections is evaluated and illustrated in 
\figref{fig:eval_num_detections} for a simulation and measurement with two pavement wetness levels (wet and moist). 
For both water levels, the simulation follows the upwards trend of increased detections from whirled up water with increasing velocity.
The detection counts per velocity lie within the spread of the measurements.
Just at $100$ and $\SI{110}{\kilo\metre/\hour}$ with a fully covered pavement, the simulated numbers are noticeably lower.

\begin{figure*}[ht]
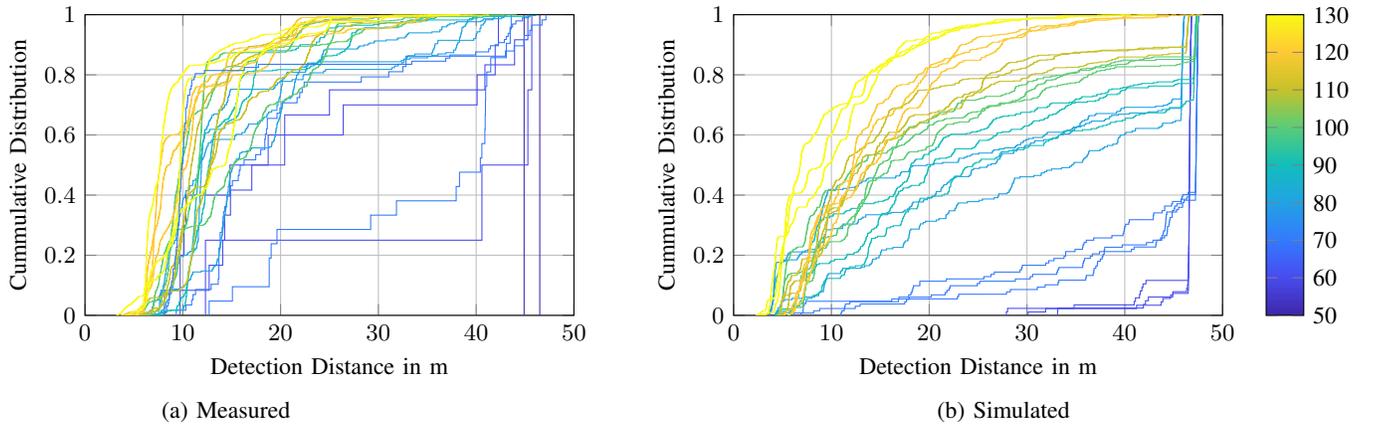

     \centering
     \begin{subfigure}[b]{0.33\textwidth}
         \centering
         \input{figures/eval_detection_distance_cdf.tikz}
         \caption{Measured}
         \label{fig:eval_detection_distance_real}
     \end{subfigure}
     \hfill
     \begin{subfigure}[b]{0.53\textwidth}
         \centering
         \input{figures/eval_detection_distance_cdf_sim.tikz}
         \caption{Simulated}
         \label{fig:eval_detection_distance_sim}
     \end{subfigure}
        \caption{Cumulative distribution of number of detections over range in spray plume from a Crafter with different speeds at $\SI{50}{\metre}$ distance. The colors denote the vehicle speed in km/h.}
        \label{fig:eval_detection_distance}
\end{figure*}

Next, the distribution over distance is evaluated.
The distance distributions of the number of detections are shown in \figref{fig:eval_detection_distance}.
The distributions also show the same trend in measurement and simulation.
Attenuation in large spay plumes lead to more detections closer to the sensor as can be seen for higher velocities.
However, for medium speed from around $\SI{80}{\kilo\metre/\hour}$ to $\SI{110}{\kilo\metre/\hour}$ detections in the simulation are visible at a greater distance, so the attenuation is not as strong as in the measurements for those speeds.

Finally, the inhomogeneity is evaluated with a clustering algorithm.
The numbers of clusters for the fully covered and the moist water level are plotted over the object speed in \figref{fig:eval_num_clusters}.
Both simulation and measurement show a similar range of cluster counts with an upwards trend over object speed suggesting a realistic modeling of the inhomogeneity.
These metrics indicate that the number of detections produced by spray, distance distribution and inhomogeneity of the simulation are within the range of the measurements.

\subsection{Object Detection Results}

\begin{table*}
\caption{Comparison of adverse weather simulation methods for two different object detectors trained on the \gls{stf} dataset. Average precision (AP) is reported for three different test sets - the spray testset with wet roads, with artificially watered roads and the \gls{stf} clear weather test set. The simulation method "spray" is the presented model in combination with the wet ground model from \cite{snow_hahner_2022}, "clear" represents the baseline without augmentation. The methods "rain", "fog", and "DROR" are competing methods.}
\centering
\setlength\tabcolsep{3pt}
\begin{tabular*}{\linewidth}{l @{\extracolsep{\fill}} l rrr | rrr | rrr }
\textbf{Detection}   &  \textbf{Simulation} &  \multicolumn{3}{c|}{\textbf{wet road}} &  \multicolumn{3}{c|}{\textbf{watered road}} & \multicolumn{3}{c}{\textbf{clear}}  \\
\textbf{method}      & \textbf{method}      & \notsotiny \textbf{0-50m} & \notsotiny 0-30m     & \notsotiny 30-50m    & \notsotiny \textbf{0-50m} & \notsotiny 0-30m     & \notsotiny 30-50m    & \notsotiny \textbf{0-50m} & \notsotiny 0-30m     & \notsotiny 30-50m     \\
\hline \hline \noalign{\vskip 1mm}
                     & spray                  & \textbf{75.36}       & \textbf{81.74}       & \textbf{67.02}       & \textbf{59.41}       & \textbf{71.83}       & \textbf{53.18}       & \underline{58.90}    & 71.69                & \textbf{44.06}        \\
                     & clear                & 57.97                & 69.75                & 42.11                & 51.34                & 66.72                & 41.63                & 58.23                & 71.33                & 42.54                 \\
                     PV-RCNN \cite{shi_pvrcnn} & rain \cite{kilic_lidar_2021}                 & \underline{67.71}    & \underline{75.82}    & \underline{55.05}    & \underline{56.18}    & 69.63                & \underline{49.16}    & \textbf{59.23}       & \textbf{71.77}       & \underline{43.88}     \\
                     & fog \cite{hahner_fog_2021}                  & 43.32                & 59.07                & 21.25                & 47.89                & 66.58                & 37.52                & 58.59                & \underline{71.73}    & 42.40                 \\
                     & DROR \cite{charron_-noising_2018}            & 52.45                & 68.93                & 35.89                & 55.54                & \underline{70.05}    & 47.93                & 54.69                & 67.26                & 39.33                 \\
\hline \noalign{\vskip 1mm}
                     & spray                  & \textbf{61.66}       & 63.27                & \textbf{58.64}       & \textbf{60.48}       & \underline{72.64}    & \textbf{53.18}       & 57.58                & 71.65                & 40.92                 \\
                     & clear                & \underline{61.04}    & \underline{63.42}    & \underline{57.56}    & 56.20                & 67.44                & 49.57                & \textbf{58.33}       & \underline{71.81}    & \textbf{41.65}        \\
                     Voxel-RCNN \cite{deng_voxel} & rain \cite{kilic_lidar_2021}                 & 54.79                & 56.37                & 52.48                & \underline{58.98}    & 71.15                & \underline{52.44}    & 57.66                & 70.84                & \underline{41.58}     \\
                     & fog \cite{hahner_fog_2021}                 & 56.15                & \textbf{64.09}       & 41.65                & 55.18                & \textbf{72.90}       & 44.61                & \underline{58.16}    & \textbf{72.38}       & 40.64                 \\
                     & DROR \cite{charron_-noising_2018}            & 49.41                & 59.79                & 32.24                & 54.66                & 69.32                & 46.79                & 53.69                & 68.61                & 36.91                 \\
\hline \noalign{\vskip 1mm}
\end{tabular*}
\label{table:results}
\end{table*}

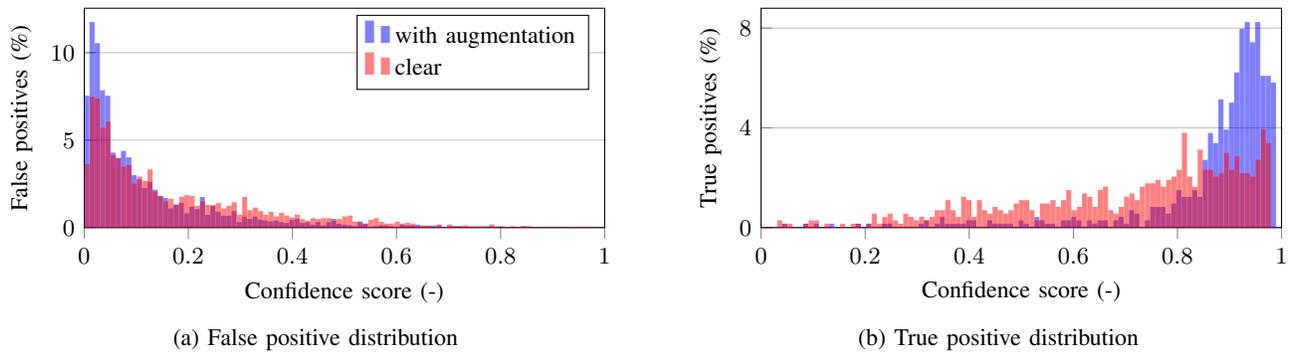
\begin{figure*}[t]
     \centering
     \begin{subfigure}[!b]{\columnwidth}
         \centering
         \begin{tikzpicture}
\definecolor{blue}{rgb}{0,0,1}
\definecolor{red}{rgb}{1,0,0}
\pgfplotsset{
    width=8.5cm,
    height=4.5cm,
}
\begin{axis}[
    ybar = -0.049cm,
    ytick={0, 5, 10},
    bar width = 1.5pt,
    enlarge y limits = {value = .07, upper},
    enlarge x limits = {abs = 0.005},
    ymajorgrids=true,
    label style={font=\small},
    legend style={font=\small},
    legend ={font=\small},
    legend pos=north east,
    legend cell align={left},
    tick pos=left,
    tick label style={font=\small},
    xlabel=Confidence score (-),
    ylabel=False positives (\%),
]
\addplot[style={fill=blue, draw=blue, mark=none}, opacity=0.5] coordinates {
    (0.005050505050505051,7.5034738304770725)
    (0.015151515151515152,11.718388142658439)
    (0.025252525252525256,10.514126910606446)
    (0.03535353535353536,7.8276980083369585)
    (0.045454545454545456,7.503473830476848)
    (0.05555555555555556,4.261232051875741)
    (0.06565656565656566,3.93700787401563)
    (0.07575757575757577,4.353867531264344)
    (0.08585858585858586,3.9833256137099315)
    (0.09595959595959597,2.964335340435298)
    (0.10606060606060608,2.7327466419637902)
    (0.11616161616161616,2.2232515053264734)
    (0.12626262626262627,2.5937934228808857)
    (0.13636363636363638,2.1306160259378704)
    (0.14646464646464646,1.7600741083834581)
    (0.15656565656565657,1.667438628994855)
    (0.16666666666666669,1.0653080129689352)
    (0.1767676767676768,1.2968967114404428)
    (0.18686868686868688,1.3895321908290459)
    (0.196969696969697,0.787401574803126)
    (0.20707070707070707,1.1579434923575382)
    (0.21717171717171718,1.0189902732746337)
    (0.2272727272727273,1.7137563686891566)
    (0.2373737373737374,0.6947660954145229)
    (0.2474747474747475,1.2042612320518398)
    (0.25757575757575757,0.8800370541917291)
    (0.26767676767676774,0.6484483557202214)
    (0.2777777777777778,0.6484483557202214)
    (0.2878787878787879,0.9263547938860306)
    (0.297979797979798,0.2779064381658092)
    (0.30808080808080807,0.6021306160259199)
    (0.31818181818181823,0.4631773969430153)
    (0.3282828282828283,0.6021306160259199)
    (0.33838383838383845,0.41685965724871377)
    (0.3484848484848485,0.37054191755441224)
    (0.3585858585858586,0.3242241778601107)
    (0.36868686868686873,0.37054191755441224)
    (0.3787878787878788,0.2779064381658092)
    (0.38888888888888895,0.23158869847150765)
    (0.398989898989899,0.41685965724871377)
    (0.40909090909090917,0.4631773969430153)
    (0.4191919191919192,0.23158869847150765)
    (0.4292929292929293,0.2779064381658092)
    (0.43939393939393945,0.1389532190829046)
    (0.4494949494949495,0.2779064381658092)
    (0.45959595959595967,0.09263547938860306)
    (0.4696969696969697,0.2779064381658092)
    (0.47979797979797983,0.4631773969430153)
    (0.48989898989898994,0.18527095877720612)
    (0.5,0.1389532190829046)
    (0.5101010101010102,0.09263547938860306)
    (0.5202020202020203,0.04631773969430153)
    (0.5303030303030303,0.3242241778601107)
    (0.5404040404040404,0.18527095877720612)
    (0.5505050505050506,0.0)
    (0.5606060606060607,0.04631773969430153)
    (0.5707070707070707,0.09263547938860306)
    (0.5808080808080809,0.09263547938860306)
    (0.5909090909090909,0.04631773969430153)
    (0.601010101010101,0.0)
    (0.6111111111111112,0.1389532190829046)
    (0.6212121212121213,0.09263547938860306)
    (0.6313131313131314,0.09263547938860306)
    (0.6414141414141414,0.1389532190829046)
    (0.6515151515151516,0.04631773969430153)
    (0.6616161616161617,0.04631773969430153)
    (0.6717171717171717,0.09263547938860306)
    (0.6818181818181819,0.09263547938860306)
    (0.691919191919192,0.0)
    (0.7020202020202021,0.04631773969430153)
    (0.7121212121212122,0.04631773969430153)
    (0.7222222222222223,0.04631773969430153)
    (0.7323232323232324,0.0)
    (0.7424242424242424,0.0)
    (0.7525252525252526,0.0)
    (0.7626262626262628,0.0)
    (0.7727272727272727,0.04631773969430153)
    (0.7828282828282829,0.04631773969430153)
    (0.792929292929293,0.0)
    (0.8030303030303031,0.04631773969430153)
    (0.8131313131313131,0.0)
    (0.8232323232323233,0.0)
    (0.8333333333333335,0.0)
    (0.8434343434343434,0.04631773969430153)
    (0.8535353535353536,0.0)
    (0.8636363636363638,0.0)
    (0.8737373737373738,0.0)
    (0.8838383838383839,0.0)
    (0.893939393939394,0.0)
    (0.9040404040404042,0.0)
    (0.9141414141414141,0.0)
    (0.9242424242424243,0.0)
    (0.9343434343434345,0.0)
    (0.9444444444444445,0.0)
    (0.9545454545454546,0.0)
    (0.9646464646464648,0.0)
    (0.9747474747474748,0.0)
    (0.9848484848484849,0.0)
    (0.994949494949495,0.0)
};
\addplot[style={fill=red, draw=red, mark=none}, opacity=0.5] coordinates {
    (0.005050505050505051,3.5967578520770025)
    (0.015151515151515152,7.446808510638331)
    (0.025252525252525256,7.345491388044646)
    (0.03535353535353536,5.6737588652482795)
    (0.045454545454545456,6.028368794326017)
    (0.05555555555555556,4.1033434650454055)
    (0.06565656565656566,3.9513677811548353)
    (0.07575757575757577,3.4447821681862667)
    (0.08585858585858586,3.5460992907799804)
    (0.09595959595959597,2.4822695035459863)
    (0.10606060606060608,2.887537993920841)
    (0.11616161616161616,2.634245187436557)
    (0.12626262626262627,3.292806484295696)
    (0.13636363636363638,2.0263424518742745)
    (0.14646464646464646,1.7730496453899902)
    (0.15656565656565657,1.469098277608849)
    (0.16666666666666669,1.6210739614994196)
    (0.1767676767676768,1.2664640324214216)
    (0.18686868686868688,1.7223910840931334)
    (0.196969696969697,1.823708206686847)
    (0.20707070707070707,1.7730496453899902)
    (0.21717171717171718,1.2158054711245647)
    (0.2272727272727273,1.469098277608849)
    (0.2373737373737374,1.2664640324214216)
    (0.2474747474747475,1.2664640324214216)
    (0.25757575757575757,1.3677811550151353)
    (0.26767676767676774,1.0638297872339941)
    (0.2777777777777778,1.2158054711245647)
    (0.2878787878787879,1.4184397163119922)
    (0.297979797979798,0.7092198581559961)
    (0.30808080808080807,1.7223910840931334)
    (0.31818181818181823,0.9625126646402804)
    (0.3282828282828283,1.114488348530851)
    (0.33838383838383845,0.9118541033434235)
    (0.3484848484848485,0.7092198581559961)
    (0.3585858585858586,0.8611955420465667)
    (0.36868686868686873,0.6079027355622824)
    (0.3787878787878788,0.8105369807497098)
    (0.38888888888888895,0.6585612968591392)
    (0.398989898989899,0.5572441742654255)
    (0.40909090909090917,0.7092198581559961)
    (0.4191919191919192,0.45592705167171177)
    (0.4292929292929293,0.35460992907799804)
    (0.43939393939393945,0.5065856129685686)
    (0.4494949494949495,0.45592705167171177)
    (0.45959595959595967,0.5065856129685686)
    (0.4696969696969697,0.5065856129685686)
    (0.47979797979797983,0.35460992907799804)
    (0.48989898989898994,0.45592705167171177)
    (0.5,0.6079027355622824)
    (0.5101010101010102,0.6585612968591392)
    (0.5202020202020203,0.3039513677811412)
    (0.5303030303030303,0.2532928064842843)
    (0.5404040404040404,0.20263424518742745)
    (0.5505050505050506,0.4052684903748549)
    (0.5606060606060607,0.5065856129685686)
    (0.5707070707070707,0.3039513677811412)
    (0.5808080808080809,0.20263424518742745)
    (0.5909090909090909,0.20263424518742745)
    (0.601010101010101,0.3039513677811412)
    (0.6111111111111112,0.20263424518742745)
    (0.6212121212121213,0.2532928064842843)
    (0.6313131313131314,0.20263424518742745)
    (0.6414141414141414,0.05065856129685686)
    (0.6515151515151516,0.10131712259371373)
    (0.6616161616161617,0.10131712259371373)
    (0.6717171717171717,0.0)
    (0.6818181818181819,0.1519756838905706)
    (0.691919191919192,0.0)
    (0.7020202020202021,0.1519756838905706)
    (0.7121212121212122,0.05065856129685686)
    (0.7222222222222223,0.05065856129685686)
    (0.7323232323232324,0.10131712259371373)
    (0.7424242424242424,0.05065856129685686)
    (0.7525252525252526,0.05065856129685686)
    (0.7626262626262628,0.05065856129685686)
    (0.7727272727272727,0.0)
    (0.7828282828282829,0.1519756838905706)
    (0.792929292929293,0.05065856129685686)
    (0.8030303030303031,0.0)
    (0.8131313131313131,0.0)
    (0.8232323232323233,0.05065856129685686)
    (0.8333333333333335,0.0)
    (0.8434343434343434,0.05065856129685686)
    (0.8535353535353536,0.05065856129685686)
    (0.8636363636363638,0.0)
    (0.8737373737373738,0.0)
    (0.8838383838383839,0.0)
    (0.893939393939394,0.0)
    (0.9040404040404042,0.0)
    (0.9141414141414141,0.0)
    (0.9242424242424243,0.0)
    (0.9343434343434345,0.0)
    (0.9444444444444445,0.0)
    (0.9545454545454546,0.0)
    (0.9646464646464648,0.0)
    (0.9747474747474748,0.0)
    (0.9848484848484849,0.0)
    (0.994949494949495,0.0)
};
\legend{with augmentation, clear}
\end{axis}
\end{tikzpicture}
         \caption{False positive distribution}
         \label{fig:false_pos}
     \end{subfigure}
     \hfil
     \begin{subfigure}[!b]{\columnwidth}
         \centering
         \begin{tikzpicture}
\definecolor{blue}{rgb}{0,0,1}
\definecolor{red}{rgb}{1,0,0}
\pgfplotsset{
    width=8.5cm,
    height=4.5cm,
}
\begin{axis}[
    ybar = -0.049cm,
    ytick={0, 4, 8},
    bar width = 1.5pt,
    enlarge y limits = {value = .07, upper},
    enlarge x limits = {abs = 0.005},
    ymajorgrids=true,
    label style={font=\small},
    legend style={font=\small},
    legend ={font=\small},
    legend pos=north east,
    legend cell align={left},
    tick pos=left,
    tick label style={font=\small},
    xlabel=Confidence score (-),
    ylabel=True positives (\%),
]
\addplot[style={fill=blue, draw=blue, mark=none}, opacity=0.5] coordinates {
    (0.005050505050505051,0.0)
    (0.015151515151515152,0.0)
    (0.025252525252525256,0.0)
    (0.03535353535353536,0.0)
    (0.045454545454545456,0.13477088948787064)
    (0.05555555555555556,0.0)
    (0.06565656565656566,0.0)
    (0.07575757575757577,0.0)
    (0.08585858585858586,0.13477088948787064)
    (0.09595959595959597,0.0)
    (0.10606060606060608,0.13477088948787064)
    (0.11616161616161616,0.0)
    (0.12626262626262627,0.0)
    (0.13636363636363638,0.13477088948787064)
    (0.14646464646464646,0.0)
    (0.15656565656565657,0.0)
    (0.16666666666666669,0.0)
    (0.1767676767676768,0.0)
    (0.18686868686868688,0.13477088948787064)
    (0.196969696969697,0.0)
    (0.20707070707070707,0.13477088948787064)
    (0.21717171717171718,0.0)
    (0.2272727272727273,0.0)
    (0.2373737373737374,0.13477088948787072)
    (0.2474747474747475,0.13477088948787053)
    (0.25757575757575757,0.0)
    (0.26767676767676774,0.0)
    (0.2777777777777778,0.0)
    (0.2878787878787879,0.0)
    (0.297979797979798,0.0)
    (0.30808080808080807,0.13477088948787053)
    (0.31818181818181823,0.26954177897574105)
    (0.3282828282828283,0.0)
    (0.33838383838383845,0.13477088948787053)
    (0.3484848484848485,0.40431266846361164)
    (0.3585858585858586,0.13477088948787053)
    (0.36868686868686873,0.13477088948787053)
    (0.3787878787878788,0.13477088948787053)
    (0.38888888888888895,0.13477088948787053)
    (0.398989898989899,0.26954177897574105)
    (0.40909090909090917,0.26954177897574105)
    (0.4191919191919192,0.0)
    (0.4292929292929293,0.0)
    (0.43939393939393945,0.26954177897574144)
    (0.4494949494949495,0.1347708894878709)
    (0.45959595959595967,0.1347708894878709)
    (0.4696969696969697,0.1347708894878709)
    (0.47979797979797983,0.1347708894878709)
    (0.48989898989898994,0.0)
    (0.5,0.2695417789757418)
    (0.5101010101010102,0.1347708894878709)
    (0.5202020202020203,0.0)
    (0.5303030303030303,0.4043126684636127)
    (0.5404040404040404,0.2695417789757418)
    (0.5505050505050506,0.1347708894878709)
    (0.5606060606060607,0.1347708894878709)
    (0.5707070707070707,0.0)
    (0.5808080808080809,0.2695417789757418)
    (0.5909090909090909,0.2695417789757418)
    (0.601010101010101,0.4043126684636127)
    (0.6111111111111112,0.26954177897574105)
    (0.6212121212121213,0.0)
    (0.6313131313131314,0.2695417789757404)
    (0.6414141414141414,0.1347708894878702)
    (0.6515151515151516,0.2695417789757404)
    (0.6616161616161617,0.2695417789757404)
    (0.6717171717171717,0.0)
    (0.6818181818181819,0.2695417789757404)
    (0.691919191919192,0.4043126684636106)
    (0.7020202020202021,0.1347708894878702)
    (0.7121212121212122,0.673854447439351)
    (0.7222222222222223,0.5390835579514808)
    (0.7323232323232324,0.0)
    (0.7424242424242424,0.2695417789757404)
    (0.7525252525252526,0.8086253369272212)
    (0.7626262626262628,0.8086253369272212)
    (0.7727272727272727,0.8086253369272212)
    (0.7828282828282829,0.5390835579514808)
    (0.792929292929293,0.9433962264150914)
    (0.8030303030303031,1.4824797843665722)
    (0.8131313131313131,1.2129380053908316)
    (0.8232323232323233,1.2129380053908316)
    (0.8333333333333335,1.4824797843665722)
    (0.8434343434343434,1.2129380053908316)
    (0.8535353535353536,2.695417789757404)
    (0.8636363636363638,3.7735849056604014)
    (0.8737373737373738,3.3692722371968244)
    (0.8838383838383839,5.121293800539172)
    (0.893939393939394,3.9083557951483163)
    (0.9040404040404042,4.9865229110513)
    (0.9141414141414141,6.19946091644214)
    (0.9242424242424243,7.951482479784177)
    (0.9343434343434345,8.221024258759913)
    (0.9444444444444445,7.4123989218327075)
    (0.9545454545454546,8.221024258759913)
    (0.9646464646464648,6.064690026954034)
    (0.9747474747474748,6.064690026954034)
    (0.9848484848484849,5.7951482479782985)
    (0.994949494949495,0.0)
};
\addplot[style={fill=red, draw=red, mark=none}, opacity=0.5] coordinates {
    (0.005050505050505051,0.0)
    (0.015151515151515152,0.0)
    (0.025252525252525256,0.0)
    (0.03535353535353536,0.2699055330634278)
    (0.045454545454545456,0.13495276653171392)
    (0.05555555555555556,0.13495276653171387)
    (0.06565656565656566,0.0)
    (0.07575757575757577,0.0)
    (0.08585858585858586,0.13495276653171387)
    (0.09595959595959597,0.2699055330634279)
    (0.10606060606060608,0.2699055330634279)
    (0.11616161616161616,0.0)
    (0.12626262626262627,0.13495276653171395)
    (0.13636363636363638,0.0)
    (0.14646464646464646,0.0)
    (0.15656565656565657,0.13495276653171395)
    (0.16666666666666669,0.0)
    (0.1767676767676768,0.13495276653171395)
    (0.18686868686868688,0.13495276653171379)
    (0.196969696969697,0.0)
    (0.20707070707070707,0.13495276653171379)
    (0.21717171717171718,0.5398110661268551)
    (0.2272727272727273,0.13495276653171379)
    (0.2373737373737374,0.40485829959514136)
    (0.2474747474747475,0.5398110661268555)
    (0.25757575757575757,0.26990553306342757)
    (0.26767676767676774,0.13495276653171379)
    (0.2777777777777778,0.5398110661268551)
    (0.2878787878787879,0.40485829959514136)
    (0.297979797979798,0.26990553306342757)
    (0.30808080808080807,0.40485829959514136)
    (0.31818181818181823,0.26990553306342757)
    (0.3282828282828283,0.40485829959514136)
    (0.33838383838383845,0.6747638326585697)
    (0.3484848484848485,0.674763832658569)
    (0.3585858585858586,1.0796221322537103)
    (0.36868686868686873,0.674763832658569)
    (0.3787878787878788,0.40485829959514136)
    (0.38888888888888895,1.2145748987854241)
    (0.398989898989899,1.0796221322537103)
    (0.40909090909090917,0.40485829959514136)
    (0.4191919191919192,1.0796221322537214)
    (0.4292929292929293,0.6747638326585759)
    (0.43939393939393945,0.5398110661268607)
    (0.4494949494949495,0.809716599190291)
    (0.45959595959595967,0.6747638326585759)
    (0.4696969696969697,0.5398110661268607)
    (0.47979797979797983,0.6747638326585759)
    (0.48989898989898994,0.9446693657220062)
    (0.5,1.0796221322537214)
    (0.5101010101010102,1.0796221322537214)
    (0.5202020202020203,0.6747638326585759)
    (0.5303030303030303,0.26990553306343035)
    (0.5404040404040404,0.9446693657220062)
    (0.5505050505050506,0.9446693657220062)
    (0.5606060606060607,1.0796221322537214)
    (0.5707070707070707,0.8097165991902883)
    (0.5808080808080809,1.0796221322537214)
    (0.5909090909090909,1.484480431848867)
    (0.601010101010101,0.6747638326585759)
    (0.6111111111111112,0.809716599190291)
    (0.6212121212121213,1.3495276653171517)
    (0.6313131313131314,1.2145748987854366)
    (0.6414141414141414,0.809716599190291)
    (0.6515151515151516,1.484480431848867)
    (0.6616161616161617,1.619433198380582)
    (0.6717171717171717,0.6747638326585759)
    (0.6818181818181819,0.5398110661268607)
    (0.691919191919192,0.809716599190291)
    (0.7020202020202021,1.2145748987854366)
    (0.7121212121212122,1.3495276653171517)
    (0.7222222222222223,1.0796221322537214)
    (0.7323232323232324,1.619433198380582)
    (0.7424242424242424,1.7543859649122973)
    (0.7525252525252526,1.7543859649122973)
    (0.7626262626262628,1.8893387314440124)
    (0.7727272727272727,1.6194331983805765)
    (0.7828282828282829,1.8893387314440124)
    (0.792929292929293,2.0242914979757276)
    (0.8030303030303031,2.294197031039158)
    (0.8131313131313131,3.778677462888025)
    (0.8232323232323233,2.0242914979757276)
    (0.8333333333333335,1.619433198380582)
    (0.8434343434343434,3.103913630229449)
    (0.8535353535353536,2.294197031039158)
    (0.8636363636363638,2.294197031039158)
    (0.8737373737373738,2.0242914979757276)
    (0.8838383838383839,2.159244264507443)
    (0.893939393939394,2.968960863697734)
    (0.9040404040404042,2.294197031039158)
    (0.9141414141414141,2.8340080971660186)
    (0.9242424242424243,2.159244264507443)
    (0.9343434343434345,2.159244264507443)
    (0.9444444444444445,2.0242914979757276)
    (0.9545454545454546,2.6990553306343035)
    (0.9646464646464648,3.91363022941974)
    (0.9747474747474748,3.3738191632928682)
    (0.9848484848484849,0.0)
    (0.994949494949495,0.0)
};
\end{axis}
\end{tikzpicture}
         \caption{True positive distribution}
         \label{fig:true_pos}
     \end{subfigure}
        \caption{Relative distribution of false and true positives over the spray test set for the PV-RCNN detector trained with and without spray augmentation. Clearly, augmentation helps to reduce confidence in false positive detections but also increases confidence in true positive ones. \vspace{12pt}}
        \label{fig:tp_fp_histo}
\end{figure*}

\begin{figure*}[ht]
	\centering
	\includegraphics[width=\textwidth]{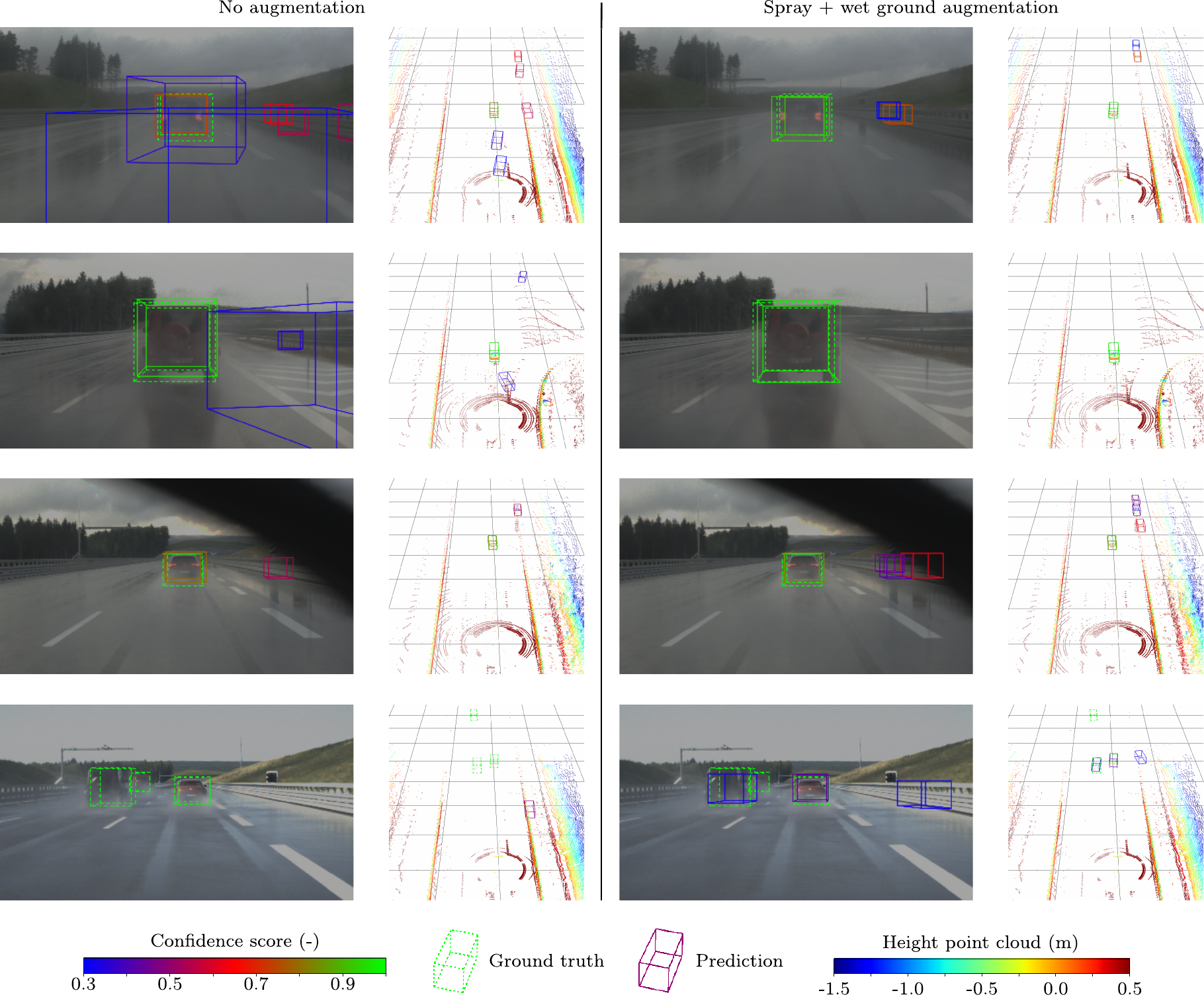}
	\caption{Qualitative comparison for the best performing PV-RCNN detector trained with augmented data with the corresponding clear weather baseline on different parts of the spray test set. The color of the bounding boxes encodes the confidence score. Ground truth boxes are additionally drawn with dashed lines.}
  \label{fig:qual_results}
\end{figure*}

Next, the model is evaluated based on how much it helps to improve object detection in real-spray conditions.
Quantitative results are presented in \tabref{table:results}.
In the three main columns, it is differentiated between the wet road spray test set, the more difficult artificially watered road test set, and the \gls{stf} clear weather test set.
The main takeaway is that augmentation with the spray simulation model improves performance on the entire evaluation range from 0-\SI{50}{\metre} for both spray test sets. 
This improvement is especially pronounced for the PV-RCNN detector.
Here, an increase of over \textbf{17\,\% AP} is achieved for wet road conditions compared to the clear weather baseline.
An improvement of \textbf{8\,\% AP} is achieved for the more difficult watered road test set.
An increase in performance on both test sets is also observed for the Voxel-RCNN detector, although not as pronounced compared to PV-RCNN.
This is consistent with the findings in \cite{snow_hahner_2022}, where the PV-RCNN detector experienced larger benefits from snow augmentation compared to Voxel-RCNN.
Both detectors use 3D voxels to encode the point cloud.
However, the PV-RCNN detector additionally extracts features from the point cloud directly and thus possibly is able to generate a larger benefit from the augmented data.

The increase in performance is further investigated in \figref{fig:tp_fp_histo}. 
Here, the relative amount of true and false positives for the PV-RCNN detector trained with augmented data is compared to the clear weather baseline.
As shown in \figref{fig:false_pos}, the increase in performance stems from better suppressing the number of false positives.
Although the spray augmentation does not lead to suppressing every false positive, it generally assigns lower confidence scores to these wrong detections. 
At the same time, the spray augmentation also helps to increase the confidence in the actual ground truth vehicles, as shown in \figref{fig:true_pos}.

When comparing to other adverse weather simulation methods, it can be seen that these models do not transfer well to spray.
On the entire evaluation range from 0-\SI{50}{\metre}, both detectors augmented with the spray simulation beat the other related adverse weather simulation model.
Only the fog simulation in the range 0-\SI{30}{\metre} seems to bring a benefit for Voxel-RCNN in spray conditions.
This might be related to the fact that water droplets from fog appear similarly to the Voxel-RCNN as smaller spray plumes.
Similarly to \cite{snow_hahner_2022}, it is found that the application of DROR does not help the performance.
The de-noising does not only remove points from spray but also valid points on vehicles.

Moreover, it is noticeable that the performance increase in spray does not result in a sacrifice in clear weather conditions.
Both object detectors trained with the spray model achieve comparable results to the respective clear weather baseline on the clear weather test set. 
Clearly, both detectors do not overfit to spray conditions.

\newpage
Finally, qualitative results are presented in \figref{fig:qual_results}.
Here, the PV-RCNN detector trained with augmentation is compared to its clear weather baseline.
In the first row, the falsely classified spray plumes are not classified as vehicles anymore.
The same can be seen for the larger G-Class in the second row which generates a larger number of spray clusters than the smaller Smart.
Several clusters together are subsequently misclassified as a vehicle in the baseline detector.
This is successfully suppressed in the detector trained with augmented data.
In the third row, it can be seen that the confidence in true positives increases significantly in the augmented detector.
The last row represents a challenging scenario form the spray test set with artificially watered road and three cars.
Here, the detector augmented with the spray simulation model is able to detect two of the three cars.
The clear weather baseline however misses all three detections.
\section{Discussion}

The objective of this work was to implement a phenomenological simulation model for spray in lidar data, based on specifically designed experiments.
The model was used to generate augmented training data and showed significant improvement of object detection algorithms in real world spray situations.
However, there are a few points to consider for future development and further application of the model.

In the calibration data set as well as the object detection data set, there is a limited amount of objects behind the spray plume.
The impact of the spray plume on the detection intensities and detection probabilities on other objects has to be further evaluated.
A gray reference target was used in the calibration data set for a first assessment, but the impact on other materials in different distances is not covered, yet.

In some scenarios in the data, the influence of objects moving through the spray plume generated by other vehicles is visible.
Currently, the spray clusters are generated for each individual moving object in the scene, there is no interaction.
The interaction of multiple objects with the spray plumes has to be further assessed.
Currently, two different objects where recorded in the data set and modeled accordingly as a first assessment of the influence of object classes.
Further research has to be conducted with more objects and also varying parameters, such as the tires and the load.
Additionally, occlusion by objects that do not generate detections themselves, like mirroring surfaces or very dark materials is not covered.
At the moment, the presented spray simulation would generate detections behind those objects.
When using the model in an SiL simulation e.g. with CarMaker, the ray tracing results could be utilized for occlusion consideration in future work.

Once these issues are addressed, the next step is to evaluate other types of lidar sensors.
Every sensor has a different optic, receiver and processing unit and first experiments have shown major differences in spray situations.
Moreover, with higher resolutions, a problem with the spherical cluster representation might appear.
The problem can become especially pronounced in close proximity to the sensor and is to be evaluated in future work.

Additionally, the presented spray model and downstream object detection methods consider only a single echo. However, most automotive lidar sensors are able to return multiple echoes. Future works should investigate modelling several echoes with respect to spray and utilizing them for detection. For instance, if the strongest echo is coming from spray, information from the hidden object might still be present in the last echo. This would likely improve object detection.

Furthermore, not only tires on wet roads produce spray, wind shield washers for example also produce water spray.
Preliminary experiments have shown that wind shield washer spray is visible in lidar data, too.
It has to be evaluated, if this phenomenon can be simulated in a similar way to road spray.

Moreover, the auto-labelling pipeline can be extended beyond highway scenarios.
The quality of the labels in other scenarios, e.g. during urban driving, will likely suffer.
This is due to the fact that the models and covariances used in the Kalman filter are tuned for highway driving.
Furthermore, associating radar detections will be increasingly difficult in urban scenarios due to lower velocities and a larger number of road users. 
Future work should consider different models for road users and association metrics in order to be applicable to more general scenarios.

Apart from application considerations, there is also the performance to be discussed.
Currently, the model is processed sequentially on an Intel i9 CPU without any performance acceleration.
The computation time of this spray model implementation in the test scenarios with only one moving object driven at $\SI{100}{\kilo\metre/\hour}$ reaches a maximum of $\SI{70}{\milli\second}$ at the time frame with the most clusters and detections.
With two moving objects, the maximum computation time doubles.
So even without parallelization, the model shows acceptable performance results on a CPU.
Running it in parallel on a GPU promises significant acceleration.

\section{Conclusion}

In this publication, a novel modeling approach for spray in lidar data is proposed simulating clusters of detections in a spray plume generated by moving vehicles.
Alongside the model, a data set for calibration and evaluation is introduced.
The data set is recorded with varying the parameters of object velocity, object class and water film height one at a time taking three measurements for every parameter combination.
From the data set, a calibration method is derived to parameterize number of cluster, cluster sizes, movement and locations as well detection probabilities for beams intersecting the clusters.

The presented simulation model was then used to generate artificial training data to improve the robustness of lidar-based object detectors.
The evaluation was performed on a specifically collected data set featuring only scenes affected by spray.
It was shown that state-of-the-art object detectors trained with artificial spray scenes can improve performance in real-world spray scenarios by up to 17\% in average precision of object recognition.

\section{Acknowledgments}
This work received funding from \emph{\mbox{SET Level}} and \emph{\mbox{VVM}} of the \emph{\mbox{PEGASUS}} project family, promoted by the German Federal Ministry for Economic Affairs and Climate Action and \emph{\mbox{VIVID}}, promoted by the German Federal Ministry of Education and Research, based on a decision of the Deutsche Bundestag.
Furthermore, the research leading to these results is part of the AI-SEE project, which is a co-labelled PENTA and EURIPIDES2 project endorsed by EUREKA. Co-funding is provided by the following national funding authorities: Austrian Research Promotion Agency (FFG), Business Finland, Federal Ministry of Education and Research (BMBF) and National Research Council of Canada Industrial Research Assistance Program (NRC-IRAP).

\bibliographystyle{IEEEtran}
\bibliography{IEEEabrv,jsen.bib}

\begin{IEEEbiography}[{\includegraphics[width=1in,height=1.25in,clip,keepaspectratio]{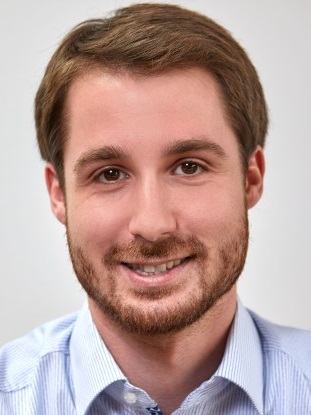}}]{Clemens Linnhoff} was born in Bochum, Germany in 1992.
He received his B.Sc. and M.Sc. degrees in mechatronics from the Technical University of Darmstadt, Germany in 2016 and 2018.
Since 2019, he has been working as Research Associate with the Institute of Automotive Engineering at TU Darmstadt.
In 2021, he co-founded the company Persival GmbH.
His research is focused on the simulation of environmental influences on radar and lidar sensors.
\end{IEEEbiography}

\begin{IEEEbiography}[{\includegraphics[width=1in,height=1.25in,clip,keepaspectratio]{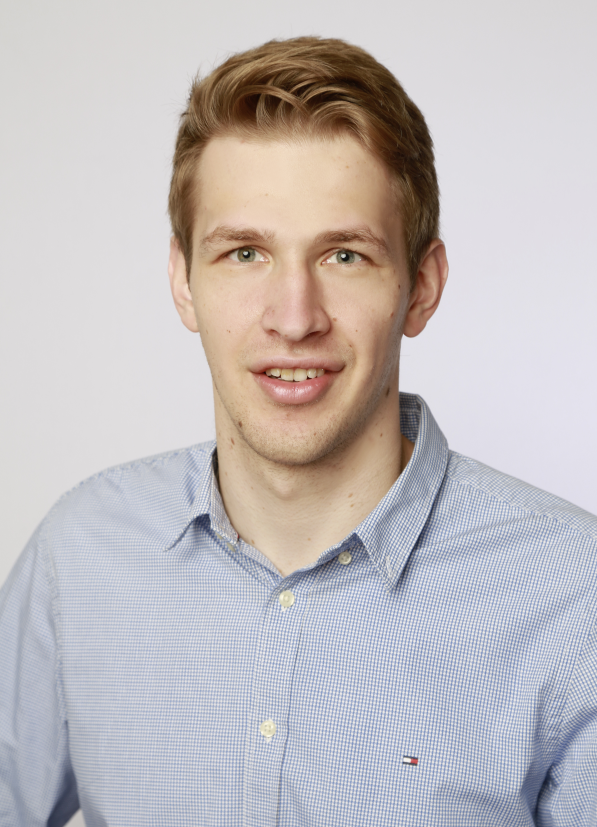}}]{Dominik Scheuble} was born in Boeblingen, Germany in 1996.
He received his M.Sc. degree in mechanical engineering from the University of Stuttgart, Germany in 2021.
Since 2021, he is working as a doctoral candidate with the Mercedes-Benz AG..
His research is focused on improving object detection for autonomous driving in adverse weather scenarios.
\end{IEEEbiography}

\begin{IEEEbiography}[{\includegraphics[width=1in,height=1.25in,clip,keepaspectratio]{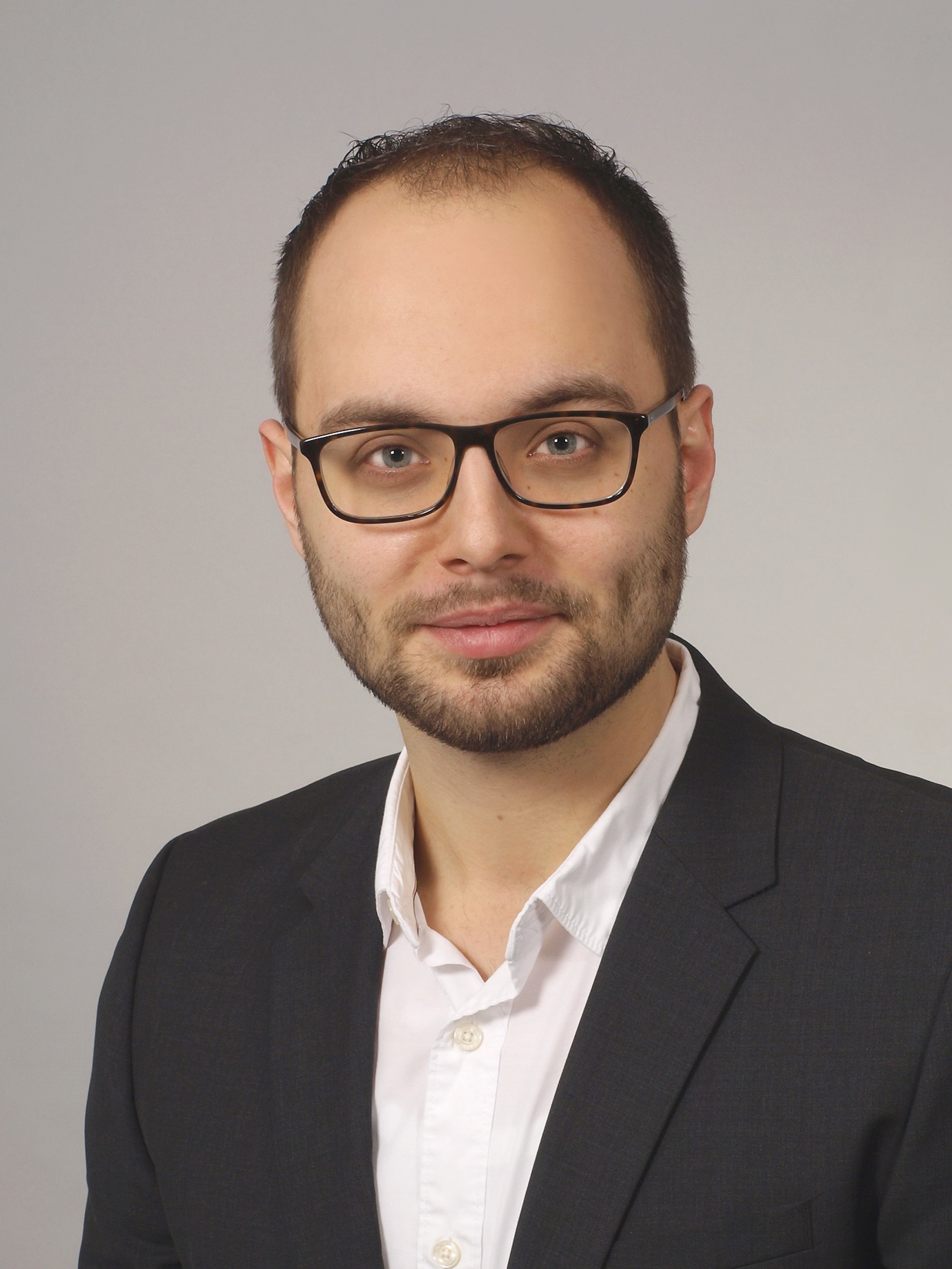}}]{Mario Bijelic} was born in Frankfurt, Germany in 1992. He received his B.Sc. and M.Sc. degrees in physics from the Johann Wolfgang Goethe University, Germany in 2014 and 2016. Since 2021, he is a Postdoctoral Researcher at Princeton University after obtaining his PhD from Ulm University. His research is focused on enabling super human vision for autonomous driving machines by applying novel hardware and computer vision algorithms overcoming todays restrictions.
\end{IEEEbiography}

\begin{IEEEbiography}[{\includegraphics[width=1in,height=1.25in,clip,keepaspectratio]{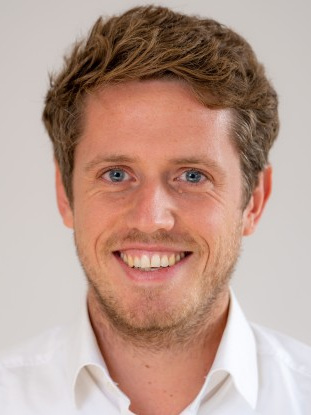}}]{Lukas Elster} was born in Fulda, Germany in 1993.
He received his B.Sc. and M.Sc. degrees in mechanical engineering from TU Darmstadt, Germany in 2016 and 2020.
Since 2020, he has been working as Research Associate with the Institute of Automotive Engineering at the TU Darmstadt.
His research is focused on the effects of radar and lidar sensors in complex surroundings.
\end{IEEEbiography}

\begin{IEEEbiography}[{\includegraphics[width=1in,height=1.25in,clip,keepaspectratio]{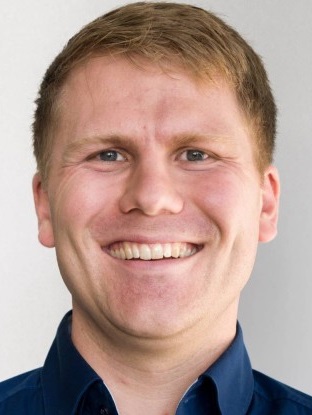}}]{Philipp Rosenberger} was born in Alzenau, Germany in 1989.
He received his B.Sc. and M.Sc. degrees in mechatronics from TU Darmstadt, Germany in 2013 and 2016.
Since 2016, he has been working as Research Associate with the Institute of Automotive Engineering at TU Darmstadt.
In 2021, he co-founded the company Persival GmbH.
He focuses on perception sensor model validation, especially on experiment design and metrics.
\end{IEEEbiography}

\begin{IEEEbiography}[{\includegraphics[width=1in,height=1.25in,clip,keepaspectratio]{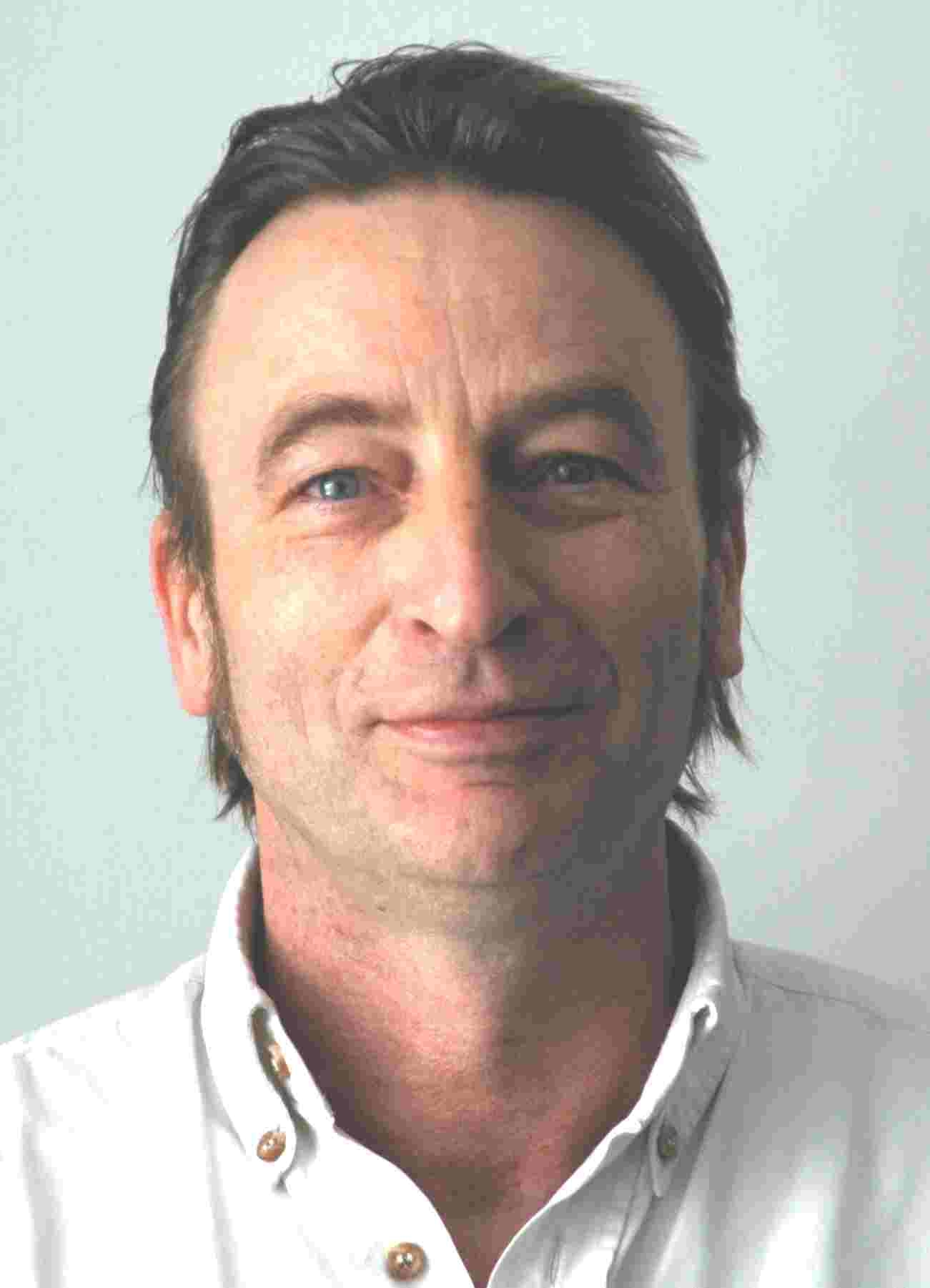}}]{Werner Ritter} received his PhD from the University of Koblenz in Computer Science in 1996. 
He has been with Daimler AG / Mercedes-Benz AG in the driver assistance systems division since 1988.
Since 2012 he is responsible for research activities in the field of perception in adverse weather. In this context, he has worked on numerous publicly funded projects, most recently projects such as RobustSENSE, DENSE, and AI-SEE.
\end{IEEEbiography}

\begin{IEEEbiography}[{\includegraphics[width=1in,height=1.25in,clip,keepaspectratio]{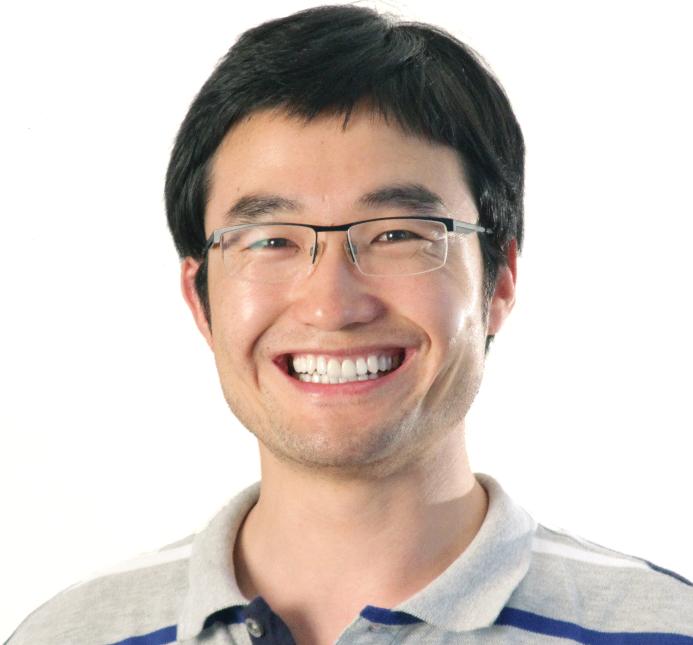}}]{Dengxin Dai} is a Senior Researcher at MPI for Informatics, heading the research group Vision for Autonomous Systems. In 2016, he obtained his Ph.D. in Computer Vision at ETH Zurich. He is a member of the ELLIS Society, Editor of IJCV, and Area Chair of CVPR 2021, CVPR 2022, and ECCV 2022. In 2021, he has received the Golden Owl Award with ETH Zurich for exceptional teaching. His research interests lie in autonomous driving, robust perception in adverse conditions, sensor fusion, multi-task learning, and object recognition in the low data regime.
\end{IEEEbiography}

\begin{IEEEbiography}[{\includegraphics[width=1in,height=1.25in,clip,keepaspectratio]{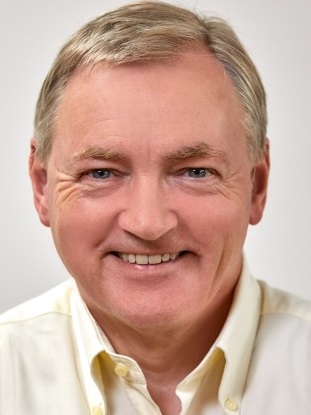}}]{Hermann Winner} was born in  Bersenbrück, Germany. After receiving his Ph.D. in physics, he began working at Robert Bosch GmbH in 1987 focusing on the predevelopment of “by-wire” technology and adaptive cruise control (ACC). Beginning in 1995, he led the series development of ACC to the start of production. 
Since 2002, he has been pursuing the research of automated driving topics as professor of Automotive Engineering at TU Darmstadt.
\end{IEEEbiography}

\end{document}